\documentclass[10pt,a4paper,english,reqno]{amsart}
\usepackage{listings,graphicx,amsmath,amsfonts,varioref,amscd,amssymb,color,bm,chemarrow,mathrsfs}
\usepackage{enumitem}
\usepackage[version=3,arrows=pgf-filled]{mhchem}
\DeclareMathOperator*{\argmin}{arg\,min}

\usepackage{subfigure,wrapfig}
\usepackage[colorinlistoftodos]{todonotes}

\usepackage[colorlinks=printing,backref]{hyperref}      

\usepackage[bbgreekl]{mathbbol}
\usepackage {amsmath, amssymb, amsthm, mathrsfs, chemarrow, amsfonts,graphicx,color,epsfig,stmaryrd,textpos,extarrows,protosem,bbm,fancybox,wasysym} 

\usepackage{mathtools}
\DeclarePairedDelimiter{\ceil}{\lceil}{\rceil}

\usepackage[T1]{fontenc}
\usepackage{ae,aecompl}
\usepackage{multicol}
\usepackage{float}
\usepackage{upref}

\newtheorem{Theorem}{Theorem}[section]
\newtheorem{Proposition}{Proposition}[section]
\newtheorem{Lemma}{Lemma}[section]
\newtheorem{Corollary}{Corollary}[section]
\newtheorem{Definition}{Definition}[section]
\newtheorem{Remark}{Remark}[section]
\newcommand{\bTheorem}[1]{\bigskip \begin{Theorem} \label{T#1}}
\newcommand{\eT}{\end{Theorem} \bigskip }
\newcommand{\bProposition}[1]{\bigskip \begin{Proposition} \label{P#1}}
\newcommand{\eP}{\end{Proposition} \bigskip }
\newcommand{\bLemma}[1]{\bigskip \begin{Lemma} \label{L#1}}
\newcommand{\eL}{\end{Lemma} \bigskip }
\newcommand{\bCorollary}[1]{\bigskip \begin{Corollary} \label{C#1}}
\newcommand{\eC}{\end{Corollary} \bigskip }
\newcommand{\bDefinition}[1]{\bigskip \begin{Definition} \label{D#1}}
\newcommand{\eD}{\end{Definition} \bigskip}
\newcommand{\bRemark}[1]{\bigskip \begin{Remark} \label{R#1}}
\newcommand{\eR}{\end{Remark} \bigskip}
\newcommand{\bFormula}[1]{\begin{equation} \label{#1}}
\newcommand{\eF}{\end{equation}}

\definecolor{darkgreen}{rgb}{0,0.75,0}
\definecolor{darkblue}{rgb}{0,0,0.75}

\usepackage[foot]{amsaddr}

\makeatletter
\renewcommand{\email}[2][]{%
  \ifx\emails\@empty\relax\else{\g@addto@macro\emails{,\space}}\fi%
  \@ifnotempty{#1}{\g@addto@macro\emails{\textrm{(#1)}\space}}%
  \g@addto@macro\emails{#2}%
}

\title[Solving Differential Equations using Deep Neural Networks]{Solving irregular and data-enriched differential equations using deep neural networks}

\author[C. Michoski]{Craig Michoski}
\address[Craig Michoski, Todd Oliver]{Oden Institute for Computational Engineering and Sciences, University of Texas at Austin, Austin,TX 78712}
\email[Craig Michoski]{michoski@ices.utexas.edu}
\author[M. Milosavljevi\'c]{Milo\v s Milosavljevi\'c }
\address[Milo\v s Milosavljevi\'c]
{Department of Astronomy, The University of Texas at Austin, Austin, TX 78712}
\author[T. Oliver]{Todd Oliver}
\author[D. Hatch]{David Hatch}
\address[David Hatch]
{Institute for Fusion Studies, University of Texas at Austin, Austin,TX 78712} 


\date{}

\begin{document}

\maketitle
\vspace{-20pt}
\begin{abstract}
 Recent work has introduced a simple numerical method for solving partial differential equations (PDEs) with deep neural networks (DNNs).  This paper reviews and extends the method while applying it to analyze one of the most fundamental features in numerical PDEs and nonlinear analysis: irregular solutions.  First, the Sod shock tube solution to compressible Euler equations is discussed, analyzed, and then compared to conventional finite element and finite volume methods.  These methods are extended to consider performance improvements and simultaneous parameter space exploration.  Next, a shock solution to compressible magnetohydrodynamics (MHD) is solved for, and used in a scenario where experimental data is utilized to enhance a PDE system that is \emph{a priori} insufficient to validate against the observed/experimental data.   This is accomplished by enriching the model PDE system with source terms and using supervised training on synthetic experimental data.  The resulting DNN framework for PDEs seems to demonstrate almost fantastical ease of system prototyping, natural integration of large data sets (be they synthetic or experimental), all while simultaneously enabling single-pass exploration of the entire parameter space.
\end{abstract}


\section{Introduction}

Solving differential equations with numerical optimization, specifically with minimization of the equation residual, is not a new idea.  Optimization-based methods have been popular for some time in the form of least-squares finite element methods (LSFEM) \cite{MR1042906,MR2490235,MR0443377}, and element-free Galerkin methods \cite{MR1256818,MR3413894}, where even for mesh-free formulations, theoretical convergence criteria have been examined \cite{MR2249154}.   Nevertheless, recent work by Han et al. \cite{Han8505}, Berg and Nystr{\"{o}}m \cite{BergN18},  Sirignano and Spiliopoulos \cite{SIRIGNANO20181339}, Raissi et al. \cite{RAISSI2019686}, and Xu and Darve \cite{1901.07758} has shown that remarkably simple implementations of deep neural networks (DNNs) can also solve relatively diverse differential equations by utilizing very similar residual-minimizing formulations but with the DNNs replacing the usual finite element discretization strategies. From a method development point of view, further demonstrating the capability of DNNs to solve differential equations is of intrinsic interest.  

To explore how efficient, reliable, robust, and generalizable DNN-based solutions are will undoubtedly require careful and prolonged study.  Here, we offer a practical starting point by asking how well, and in what ways, can DNNs manage one of the cleanest, simplest, and most ubiquitous irregularities observed in differential systems, namely shock fronts.  Shock fronts provide a quintessential test bed for a numerical method in that they contain an isolated analytically non-differentiable feature that propagates through the differential system.  The feature reduces the local regularity of the solution both in space and time and leads to numerical representations that must accommodate local first-order analytic discontinuities
while still maintaining some concept of numerical stability and robustness.   Indeed the way a numerical method responds to a local shock front tends to reveal the fundamental signature response that the numerical method has to local irregularities in the solution, which might be thought of as providing an indicator by which to gauge the method's applicability.

A tremendous amount of work spanning pure mathematical analysis \cite{MR0350216,MR688146}, numerical analysis \cite{ISI:000182615000001, ISI:000221833000006}, physics \cite{ ISI:A1992HY14300014,ISI:000222531400088}, etc., has been done on differential systems that demonstrate shock-like behavior.   
Shock-like numerical behavior is deeply rooted in discrete function representation and approximation theory, and further in functional and discrete functional analysis, specifically in what it means to locally and globally approximate a member of a function space \cite{MR2372235, MR1669959, MR628971, MR1261635}.

Along these lines, the first half of this paper is dedicated to exploring the behavior of the conventional Sod shock tube solution to the compressible Euler equations of gas dynamics \cite{SOD19781}.  We use the shock tube solution as a setting in which to explore what is meant by a DNN solution to a system of PDEs, what is meant by a regular (and irregular) solution to a system of PDEs both analytically as well as numerically, how these two frames of reference relate to each other, and where they differ.  Then we discuss in some detail how DNN solutions compare and relate to more traditional and conventional ways of solving systems of PDEs and how the validation of DNN-based methods is slightly different than, say, typical numerical convergence analysis.

The second half of the paper is focused more directly on how to improve the performance of DNN solutions in the context of irregular PDE solvers, and how to capitalize on the operational details of DNNs solvers to perform complete and simultaneous parameter space exploration.  We also consider how to extend the framework of physics-based modeling to incorporate experimental data into the analytic workflow.  These questions have growing importance in science and engineering applications as data becomes increasingly available.  Many of the more traditional approaches, including data-informed parameter estimation methods \cite{ISI:000311688000013} and real-time filtering \cite{ISI:A1993LE24400004}, are frequently incorporated into more standard forward PDE solvers. We argue that DNN-based PDE solvers provide a natural interface between physics-based PDE solvers and advanced data analytics.  The general observation is that because DNNs are able to combine a PDE solver framework with a simple and flexible interface for optimization, it becomes natural to ask how sufficient a modeling system is at representing experimental observations and to explore the nature of the mismatch between a simulated theoretical principle and an experimentally measured phenomenon.  In this context, the DNN framework almost automatically inherits the ability to couple many practical data-driven concepts, from signal analysis, to optimal control, to data-driven PDE enrichment and discovery.

To provide some insight into the utility of the DNN framework for solving PDEs, it is instructive to discuss some of the apparent strengths and drawbacks. Three strengths of the framework are: 
\begin{enumerate}[label=(\Roman*)]
\item Phenomenal ease of prototyping PDEs, 
\item Natural incorporation of large data, 
\item Simultaneous solution over entire parameter space. 
\end{enumerate} Regarding (I), complicated systems of highly parameterized and multidimensional PDEs can be prototyped in TensorFlow or PyTorch in hundreds of lines of code, in a day or two. This might be compared to the decade-long development cycle of many legacy PDE solvers.  For (II), incorporating and then utilizing experimental data in the PDE workflow as a straightforward supervised machine learning task is remarkably rewarding and provides a painless integration with the empirical scientific method.  It is easy to see how this feature could be naturally leveraged for practical uncertainty quantification, risk assessment, data-driven exploration, optimal control, and even discovery and identification of the theoretical underpinnings of physical systems (as discussed in some detail in section \ref{Sec:enriched} below).  In (III), another powerful and practical advantage is identified, where $n$-dimensional parameter space exploration simply requires augmenting the solution domain (space-time) with additional parameter axes, $(\boldsymbol{x},t,p_1, \ldots, p_n)$, and then optimizing the DNN to solve the PDEs as a function of the parameters as inputs.  The input space augmentation adds little  algorithmic or computational complexity over solving on the space-time domain $(\boldsymbol{x},t)$ at a single parameter point $(p_1,\ldots, p_n)$ and is drastically simpler than exploring parameter space points sequentially.  

Some of the more immediate drawbacks the DNN-based approach to solving PDEs include: 
\begin{enumerate}[label=(\roman*)]
    \item Absence of theoretical convergence guarantees for the non-convex PDE residual minimization,
    \item Slower overall run time per forward solve,
    \item Weaker theoretical grounding of the method in PDE analysis.
\end{enumerate}
We touch on these weaknesses throughout the paper, but briefly, (i) indicates the challenge posed by understanding convergence criteria in convex optimization, where solutions may get trapped in local minima.  The concern of (ii) is substantially subtler than it appears at first glance, and strongly depends on many aspects of the chosen neural network architecture, how optimized the hyperparameters are, what the ultimate goal of the simulation is, etc.  Finally, (iii) merely highlights that DNNs have only recently been seriously considered for PDE solvers, and thus, remain largely theoretically untouched.

The paper is organized as follows: Section \ref{Sec2} shows how a simple DNN can be used to solve a nonlinear system of PDEs.  Section \ref{Sec3} proceeds with a discussion about the relationship between different notions of solutions to PDEs and attempts to provide some context in which to understand where DNN solutions can be placed relative to solutions provided by more conventional numerical methods. Section \ref{Sec4} is a demonstration of how the DNN solution behaves given unbounded gradients.
Section \ref{Sec5}  explores solutions to the Sod shock tube problem and compare aspects of the DNN method to more standard finite element and finite volume methods. Section \ref{Sec6} discusses a method of dissipative annealing to improve (in a specific sense) the rate of convergence relative to the number of iterations required in the descent algorithm, and then how to use the DNN framework to recover simultaneous parameter scans..  In section \ref{sec:LSTM} the modified network with residual connections and multiplicative gates from  \cite{SIRIGNANO20181339}, which resembles the ``highway network'' and ``long short term memory'' (LSTM) architectures,  is compared to the more standard DNN model in the specific context of the Sod shock tube.  Finally in section \ref{Sec:enriched}, the concept of data-enriched PDEs is presented through the lens of a hypothetical experimental scenario, where the experimenter finds herself in the situation in which the anticipated physical model of the system does not adequately validate against the experimental data.   The accompanying code to this manuscript will be made available on github upon publication.

\section{Solving PDEs using Deep Neural Networks}\label{Sec2}

In this paper we are interested in a broad system of $n$-coupled potentially nonlinear PDEs that can be written in terms of the initial-boundary value problem, for $i=1,\ldots,n$: \begin{equation}\begin{aligned} \label{base} \partial_{t}u_i + \mathcal{N}_i(\boldsymbol{u};\boldsymbol{p}) & = 0, \quad \mathrm{for} \ (\boldsymbol{x},t,\boldsymbol{p})\in \Omega\times [0,T_s] \times \Omega_{\boldsymbol{p}},\\ u_i|_{t=0} & = u_{i,0}(\boldsymbol{x},\boldsymbol{p}), \quad \mathrm{for} \ (\boldsymbol{x},\boldsymbol{p})\in\Omega\times\Omega_{\boldsymbol{p}}, \\ \mathcal{B}_i(\boldsymbol{u},\boldsymbol{p}) & = 0, \quad \mathrm{for} \  (\boldsymbol{x},t,\boldsymbol{p})\in \partial\Omega\times [0,T_s]\times\Omega_p,\end{aligned}\end{equation} over the domain $\Omega\subset \mathbb{R}^{d}$ with boundary $\partial\Omega$, and the $m$ dimensional parameter domain $\boldsymbol{p}\in\Omega_{\boldsymbol{p}}$ given $m$ parameters $\boldsymbol{p}=(p_1,\ldots ,p_m)$, where $\boldsymbol{u}=\boldsymbol{u}(\boldsymbol{x},t,\boldsymbol{p})$ is the $n$ dimensional solution vector with $i$th scalar-valued component $u_i=u_i(\boldsymbol{x},t,\boldsymbol{p})$, $T_s$ is the duration in time, $\mathcal{N}_i$ is generally a nonlinear differential operator, with arbitrary boundary conditions expressed through the component-wise operator $\mathcal{B}_i(\boldsymbol{u},\boldsymbol{p})$.

Here we are interested in approximating the solutions to (\ref{base}) with DNNs.  A feed-forward network can be described in terms of the input $y\in\mathbb{R}^{d_{\mathrm{in}}}$ for $y=(\boldsymbol{x},t,\boldsymbol{p})$, the output $z^{L}\in\mathbb{R}^{d_{\mathrm{out}}}$, and an input-to-output mapping $y \mapsto z^{L}$, where $d_{\mathrm{in}}$ and $d_{\mathrm{out}}$ are the input and output dimension.  The components of the pre- and post-activations of hidden layer $\ell$ are denoted $y_{k}^{\ell}$ and  $z_{j}^{\ell}$, respectively, where the activation function is a sufficiently differentiable function $\phi:\mathbb{R}\rightarrow\mathbb{R}$.  The $j$th component of the activations in the $\ell$th hidden layer of the network is given by
\begin{equation}\label{DNNs}z_{j}^{\ell}(y) = b_{j}^{\ell} + \sum_{k=1}^{N_{\ell}}W_{jk}^{\ell}y_{k}^{\ell}(y),\quad y_{k}^{\ell}=\phi(z_{k}^{\ell-1}(y)), 
\end{equation} 
where $N_{\ell}$ are the numbers of neurons in the hidden layers of the network, and $W_{jk}^{\ell}$ and $b_{j}^{\ell}$ are the weight and bias parameters of layer $\ell$.  These parameters make up the tunable, or ``trainable'', parameters of the network, which can be thought of as the degrees of freedom that parametrize the representation space of the DNN.  Note that in section \ref{sec:LSTM} we will consider a slightly more exotic network architecture, but for now, we concern ourselves with the standard ``multilayer perceptron'' described in (\ref{DNNs}).

In this context, the output of the neural network $z^{L}$, where the number of units $N_{\ell}$ is chosen and fixed for each layer, realizes a family of approximate solutions to (\ref{base}), such that for each $i$ component of (\ref{base}), 
\begin{equation}\label{approx}z_{i}^{L}(\boldsymbol{x},t;\boldsymbol{p},\boldsymbol{\vartheta}) \approx u_{i}(\boldsymbol{x},t,\boldsymbol{p}),\end{equation}  where we have set the parameters $\boldsymbol{\vartheta} = (W,b)$.
Similarly, activation functions $\phi$ must be chosen such that the differential operators \[\frac{\partial z_{i}^{L}}{\partial t}(\boldsymbol{x},t;\boldsymbol{p},\boldsymbol{\vartheta})\quad \mathrm{and}\quad \mathcal{N}_{i}(z^{L};\boldsymbol{p}),\] can be readily and robustly evaluated using reverse mode automatic differentiation \cite{7332676320120401}. 

We proceed by defining a slightly more general  objective function to that presented in \cite{RAISSI2019686,SIRIGNANO20181339},
\begin{equation}
\label{loss}
J_i(\boldsymbol{\vartheta}) = 
J^{\mathrm{PDE}}_i(\boldsymbol{\vartheta}) + 
J^{\mathrm{BC}}_i(\boldsymbol{\vartheta}) + 
J^{\mathrm{IC}}_i(\boldsymbol{\vartheta})
\end{equation}
where
\begin{equation}
\begin{aligned}
J^{\mathrm{PDE}}_i(\boldsymbol{\vartheta}) =
\bigg\| \frac{\partial z_{i}^{L}}{\partial t} + \mathcal{N}_{i}(z^{L};\boldsymbol{p}) \bigg\|_{\mathfrak{L}_i(\Omega\times[0,T_s]\times\Omega_{\boldsymbol{p}},\wp_{1})}, & \quad 
J^{\mathrm{BC}}_i(\boldsymbol{\vartheta}) =  \|\mathcal{B}_i(\boldsymbol{u},\boldsymbol{p})\|_{\mathfrak{L}_i(\partial\Omega\times[0,T_s]\times\Omega_{\boldsymbol{p}},\wp_{2})},\\ J^{\mathrm{IC}}_i(\boldsymbol{\vartheta}) = \|z_{i}^L(\boldsymbol{x},0;\boldsymbol{\vartheta})-u_{i,0}(\boldsymbol{x}&,\boldsymbol{\vartheta}) \|_{\mathfrak{L}_i(\Omega\times\Omega_{\boldsymbol{p}},\wp_{3})},
\end{aligned}
\end{equation}
and $\mathfrak{L}_{i}$ indicates the form of the $i$th loss and $\wp_{1,2,3}$ denotes the probability density with respect to which the expected values of the loss are calculated.  In particular, the form of the loss function can correspond to any standard norm (e.g., the discrete kernel of an $L^p$-norm), the mean square error, or specialized hybrids such as the Huber loss \cite{huber1964}.   For this chosen form $\mathscr{L}$, the corresponding loss is given by the following expected value:
\begin{equation*}
\|f(X)\|_{\mathfrak{L}(\mathcal{Y},\wp)} = \mathbb{E}_{\wp}[\mathscr{L}(f)] = \int_{\mathcal{Y}} \mathscr{L}(f) \wp(X) dX.
\end{equation*}
Letting $z^L = \{z^L_1,\ldots,z^L_n\}$, we define any particular approximate solution $\hat{z}^{L}$ to (\ref{base}) as finding a $\boldsymbol{\vartheta}$ that minimizes the loss relative to the parameter set. That is: 
\begin{equation}
\label{opt}
\hat{\boldsymbol{\vartheta}}  = \argmin_{\boldsymbol{\vartheta}}\sum_{i=1}^{n}J_i(z_i^L). 
\end{equation}  

To solve this optimization problem, one must approximate the loss function and couple this approximation with an optimization algorithm.  For instance, using stochastic gradient descent (SGD), one uses a Monte Carlo approximation of the loss coupled with gradient descent.  Thus, the loss is computed by drawing a set of samples $X_{\mathrm{batch}}$ from the joint distribution $\wp_1 \wp_2 \wp_3$ and forming the following approximation:
\begin{equation*}
    G_i(\boldsymbol{\vartheta}; X_{\mathrm{batch}}) =
    G_i^{\mathrm{PDE}}(\boldsymbol{\vartheta}; X_{\mathrm{batch}}) +
    G_i^{\mathrm{BC}}(\boldsymbol{\vartheta}; X_{\mathrm{batch}}) +
    G_i^{\mathrm{IC}}(\boldsymbol{\vartheta}; X_{\mathrm{batch}}),
\end{equation*}
where 
\begin{equation*}
\begin{aligned}
G_i^{\mathrm{PDE}}(\boldsymbol{\vartheta}; X_{\mathrm{batch}}) = 
\frac{1}{N_{\mathrm{PDE}}} \sum_{n=1}^{N_{\mathrm{PDE}}} \mathscr{L}( R_i (\mathbf{x}_n, t_n, \mathbf{p}_n) ), & \qquad
G_i^{\mathrm{BC}}(\boldsymbol{\vartheta}; X_{\mathrm{batch}}) =
\frac{1}{N_{\mathrm{BC}}} \sum_{n=1}^{N_{\mathrm{BC}}} \mathscr{L}( B_i (\mathbf{x}_n, \mathbf{p}_n) ), \\
G_i^{\mathrm{IC}}(\boldsymbol{\vartheta}; X_{\mathrm{batch}}) =
\frac{1}{N_{\mathrm{IC}}} \sum_{n=1}^{N_{\mathrm{IC}}} \mathscr{L}( z_i^L(\mathbf{x}_n, &0, \mathbf{p}_n; \boldsymbol{\vartheta}) - u_{i,0}(\mathbf{x}_n, \mathbf{p}_n) ),
\end{aligned}
\end{equation*}
and $R_i(\mathbf{x}_n, t_n, \mathbf{p}_n)$ denotes the $i$th PDE residual evaluated at the batch point $(\mathbf{x}_n, t_n, \mathbf{p}_n)$, which is drawn according to $\wp_1$, and similarly for the BC and IC terms. Then, the SGD algorithm for the $k+1$ iteration can be written as
\begin{equation}
\label{sgd}  
\boldsymbol{\vartheta}_{k+1} \leftarrow \boldsymbol{\vartheta}_{k} - \alpha_{k}\nabla_{\boldsymbol{\vartheta}}\left(\sum_{i=1}^{n} G_i(\boldsymbol{\vartheta}_k; X_{\mathrm{batch}}^k)\right),
\end{equation}
where $X_{\mathrm{batch}}^k$ denotes the batch set drawn at iteration $k$.

In practice we find the Adam optimizer \cite{1412.6980} to be the most efficient SGD-family algorithm for the cases run in this paper.  For network initialization we use the Xavier uniform initializer \cite{pmlr-v9-glorot10a}.  It is also worth noting that the hyperparameters $L$ and $N_{\ell}$ as well as the sampling distribution drawn from $\wp$ are additional hyperparameters that can be optimized for.  For simplicity, however, we do not perform hyperparameter optimization in this paper.

\section{Irregular Solutions to PDEs}\label{Sec3}

In this section we briefly review what a solution to a PDE represents.  Although most PDEs have been derived from and are studied for their ability to represent natural phenomena, at their core they remain mathematical abstractions that are only approximations to the phenomena that inspire them.  These abstractions are often derived from simple variational perturbation theories, and are subsequently celebrated for their ability to capture and predict the physical behavior of complex natural systems.  This ``capturing'' of physical systems is conventionally accomplished at the level of experimental validation, which is often many steps removed from the abstraction that the PDE itself represents.

In the study and understanding of PDEs and mathematical analysis, it is difficult to over-emphasize the importance of stability and regularity.   One of the most challenging problems in modern theoretical science and engineering is developing a strategy for solving a PDE and ultimately interpreting the solution.   In this context, irregular solutions to PDEs are easy to come by.  To account for this, convention introduces the concept of mathematically well-posed solutions (in the sense of Hadamard \cite{cite:hadamard}).  Well-posedness circumscribes what is meant, in some sense, by a ``good'' or admissible solution to a PDE. It is within this framework that the concept of classical and regular solutions, strong solutions, weak solutions, and so forth has evolved.  

A ``regular solution'' can be thought of as one where the actual PDE is satisfied in an intuitively sensible and classical way.  Namely, in a regular solution, enough derivatives exist at every point in the solution so that operations within the PDE make sense at every point in the entire domain of relevance.  A well-posed regular solution is formally defined to provably exist, be unique, and depend continuously on the data given in the problem \cite{evans10}.  In contrast to regular solutions, the concept of a weak solution was developed to substantially weaken the notion of what is meant by a solution to a PDE.  In weak solutions derivatives only exist in a ``weak'' distributional sense, and the solution is subsequently recast in terms of distributions and integral norms.

For example, in the case of a ``strong solution'' to a PDE, which is a ``weak solution'' that is bounded and stable in the $L^2$-norm and thus preserves important geometric properties, a solution may admit, e.g., pointwise discontinuities and remain an admissible solution.  Such solutions are the strongest, and/or most well-behaved of the weak solutions; still, interpreting the space of strong solutions from a physical point of view can be challenging.  Moreover, weak solutions in the sense of Hadamard are only proper solutions to a PDE if they are adjoined with a concept of stability (such as $L^p$-stability) that implies a bound in a given normed space.  This is a rather practical condition, in that a solution in a distributional sense that is not bounded in norm is too permissive to pointwise ``variation'' to be physically interpretable.

A substantial fraction of solutions arising in applied science are weak solutions that demonstrate stability in notably irregular spaces.  Even the simplest of equations can admit solutions of highly irregular character (e.g., the Poisson equation \cite{10.2307/20535956}).   In fact, in many applied areas the concept of turbulence becomes important, which from a mathematical point of view deals with PDEs in regimes where neither regularity nor stability are observed, a fact with broad conceptual ramifications \cite{ISI:000447886500004}.  This irregular and unexpected behavior in PDE analysis leads to a rather basic and inevitable question that we address in the next section: how well do our specific discrete numerical approximate solutions to certain PDEs capture their rigorously established mathematical properties? 


\subsection{Numerical solutions}

Discrete numerical systems cannot, in general, exactly replicate infinite dimensional spaces (except in an abstract limit), and thus approximate numerical solutions to PDEs remain just that: approximations to mathematical PDEs.  As a matter of practice, weak formulations are often implemented in numerical methods, but while these methods often preserve the spirit of their mathematical counterparts, they subsequently lose much of their content.  For example, solutions in discontinuous finite element methods are frequently couched in terms of $L^2$ residuals and are presented along with numerical stability results, when in effect the discrete $L^2$ spaces they are purported to represent are simply piecewise continuous polynomial spaces that are capable of representing only a tiny fraction of admissible solutions in $L^2$. 

It is then a natural question to ask, what has become of all the admissible solutions that even the most rigorous numerical methods discard by construction?  And how exactly has it been determined which solutions are to be preserved?  A pragmatist might respond with the argument that numerical methods develop, primarily, to deliver practical utility, adjoining as a central thematic pillar to their evolution the notion of ``physical relevance,'' where physical relevance is notionally defined in terms of the utility that the numerical solution provides in the elucidation of a practical observation.

While this may certainly be the case, and while we adopt the pragmatic perspective in this paper, we would like to raise a consideration to the reader: if numerical methods, particularly those driving the evaluation of predictive simulation models, are, with widespread application, utilizing numerical schemes that to some extent arbitrarily select solutions of a particular kind from the space of mathematically admissible solutions, and then use that subset of solutions as the justification upon which interpretive predictive physical understanding is derived, then how can one evaluate the predictive capability of a model without directly comparing it to observed data?

\subsection{Shocks}

Perhaps the simplest type of irregularity that confronts a numerical method is a numerical shock.  A numerical shock need not be a mathematical discontinuity, and as such it is a fundamentally different concept than both a mathematical shock and a physical shock.   One definition of a numerical shock is: the numerical response (in the form of numerical instability) of a method when the representation space that the method is constructed in does not support the function space it is interrogating.  Examples of this behavior can be found from Gibb's phenomena in spectral methods to Runge's phenomena in polynomial based methods and elsewhere.  More broadly, one can simply posit that the radius of convergence of the solution must be contained within the support of its discrete representation, or else instabilities are likely to arise \cite{doi:10.1098/rspa.1957.0078,doi:10.1137/0150091}.  As a consequence, numerical approaches to the shock problem display unique characteristics that often expose deep numerical insights into the nature of the method.  The unique signature that a numerical method displays when dispersing its instabilities throughout a solution foretells the predictive characteristics of the simulation itself.

Many successful numerical methods can be viewed as variations of general strategies with important but subtle differences in the underlying detail. Consider, for example, finite volume methods (FVM), spectral, pseudospectral, and Fourier Galerkin methods, continuous and discontinuous Galerkin finite element methods, etc.  In each of these cases, the system of PDEs is cast into a weak formulation, where the spatial derivatives are effectively transferred to analytically robust basis functions, and fluxes are recovered from  integration by parts.  For example, if we choose some adequately smooth test function $\boldsymbol{\varphi}$, multiply it by some transport equation $\partial_{t}\boldsymbol{U} + \boldsymbol{F}( \boldsymbol{U})_{x} = 0$ in a state space where $\boldsymbol{U}$ a state vector, and integrate by parts, we have an equation that can be written in the form: \begin{equation}\begin{aligned}\label{weak} & \frac{d}{dt} \int_{\Omega} \boldsymbol{\varphi}\odot \boldsymbol{U} dx + \int_{\partial\Omega} \boldsymbol{\varphi} \odot \boldsymbol{F} dS - \int_{\Omega} \boldsymbol{F} \odot \boldsymbol{\varphi}_x dx = 0, \end{aligned}\end{equation} where $\odot$ denotes componentwise multiplication.

To see the relation between distinct numerical methods, recall here that when $\boldsymbol{\varphi}$ is a high order polynomial, and the discrete domain is, for example, a single element, the recovered method is a type of spectral element method (SEM).  Whereas, when the discretization is turned into a finite element mesh $\Omega_h$ comprised of $i$ subintervals, $\Omega_{1},\ldots, \Omega_{i}$, and the basis functions are chosen to be continuous or piecewise discontinuous polynomials, we have the continuous Galerkin (CG) or discontinuous Galerkin (DG) finite element methods, respectively.  Similarly, when the elements $\Omega_{i}$ are viewed as finite volumes and the basis functions  $\boldsymbol{\varphi}$ are chosen to be piecewise constant, a finite volume method can be easily recovered.

As discussed above the weak formulation (\ref{weak}) of method should not really be viewed, in and of itself, as an advantage.  Indeed, the DNN loss functions (\ref{loss}) could equally be cast into a weak formulation (as in \cite{MR2490235} for example), paying a price in the simplicity of the method.  Similarly, by using different activation functions (e.g., $\phi$ being taken as ReLU), the universal approximation theorem can, in the limit sense, recover all Lebesgue integral function spaces \cite{NIPS2017_7203}.  We sidestep these issues, viewing them as unnecessary complications that muddy the simplicity and elegance of the DNN solution.  Instead, we choose the hyperbolic tangent $\tanh(z)$ activation function and interpret solutions to shock problems through the lens that smooth solutions are dense in $L^p$.

\subsection{Euler's equations and the Sod shock tube}

For simplicity, we start by considering the dimensionless form of the compressible Euler equations in 1D, solved over only the space-time domain $\Omega\times [0,T_{s}]$, with $\Omega = [-1,1]$ and boundary $\partial\Omega = \{-1,1\}$.  We are interested in solutions to the initial-boundary value problem: \begin{equation}\begin{aligned} \label{euler} & \partial_{t}\boldsymbol{U} + \partial_x \boldsymbol{F}(\boldsymbol{U}) = 0, \quad \boldsymbol{U}_{|t=0} = \boldsymbol{U}_{0}, \quad \boldsymbol{U}_{|x\in\partial\Omega} = \boldsymbol{U}_{\partial\Omega}, \end{aligned}\end{equation} for a state vector $\boldsymbol{U} = (\rho,\rho u, E)^{\top}$ and corresponding flux $\boldsymbol{F}(\boldsymbol{U}) = (\rho u, \rho u^2 +p, u(p+E))^{\top}$.   The state vector is comprised of the density $\rho$, velocity $u$, and total energy of the gas \[E =\frac{p}{\gamma-1} +\frac{\rho}{2}u^2,\] while the pressure-temperature relation is the single component ideal gas law $p=\rho T$.  The constant $\gamma$ denotes the adiabatic index and, unless otherwise noted, is taken throughout the remainder of the paper to be $\gamma=5/3$, while the speed of sound is $c_s=\sqrt{\gamma T}$.

To analyze the behavior of the network on a simple and classic 1D shock problem, we solve the Sod shock tube.  The initial condition is chosen to be piecewise constant: \begin{equation} \boldsymbol{U}_{0} = \begin{cases} \boldsymbol{U}_{l}, &  x <0  \\ \boldsymbol{U}_r, & x\geq 0  \end{cases} \end{equation} where the left state is defined by $\boldsymbol{U}_l = (1,0,1.5)^{\top}$ and the right by $\boldsymbol{U}_r = (0.125,0,0.15)^{\top}$.  This also defines the left and right Dirichlet boundaries $\boldsymbol{U}_{\partial\Omega} = \{\boldsymbol{U}_{l},\boldsymbol{U}_r\}$. The exact solution to this system is readily obtained by a standard textbook procedure \cite{lev02}. Except for in section \ref{Sec:enriched}, the simulation time horizon is set to $T_{s}=0.2$. 

\section{Numerical Experiments}\label{Sec4}

In this section we show the basic behavior of DNN solutions to the Sod shock tube.  In section \ref{Sec4.1} we show how the DNN solution behaves without any regularization and discuss how that compares to other numerical methods.  In section \ref{Sec4.2} we discuss how to add analytic diffusion to regularize the solution.

\subsection{An unlimited solution using DNNs}\label{Sec4.1}

The standard solution to the Sod shock tube problem for (\ref{euler}) poses real challenges to most numerical methods.  In the standard DG method, for example, when the polynomial degree is nonzero, $n>0$, the solution becomes numerically unstable without the use of some form of slope limiting \cite{ISI:000302933100009,ISI:000368167500017}.  Similarly when FVM is evaluated at higher than first order accuracy, the absence of flux-limiting leads to unstable solutions \cite{ISI:000351438000007, ISI:000249936500010}.  This behavior is also observed in spectral methods \cite{ ISI:000268295900011,ISI:A1992HJ79300004}, continuous Galerkin methods \cite{ISI:000428966600004}, and so on.

\begin{figure}[t]
\includegraphics[width=0.5\textwidth]{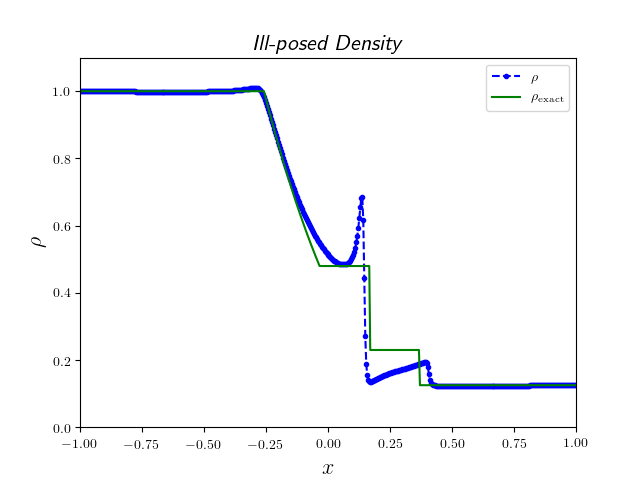}\includegraphics[width=0.5\textwidth]{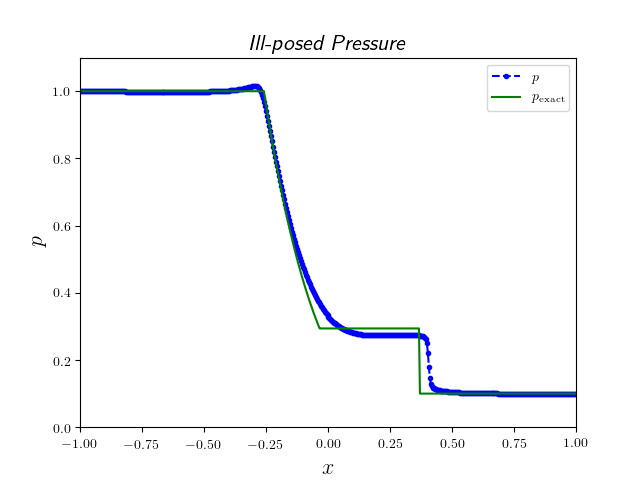} \\ \includegraphics[width=0.5\textwidth]{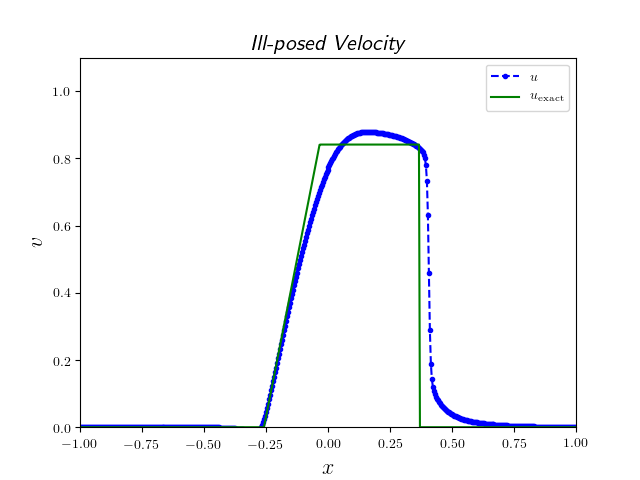}\caption{The unlimited solutions to the Sod shock tube.  The solution is fully ill-posed in the classical formulation, leading to undefined behavior (i.e., unbounded derivatives) along the shock fronts.  Nevertheless, the solution remains stable (in contrast with the unlimited versions of other algorithms), and appears relatively robust to the major features of the solution.}\label{fig:1}
\end{figure}

One of the more robust regimes for solving shock problems can be found in constant DG methods with degree $n=0$ basis functions (or equivalently, FVM methods with first order accuracy), wherein both a stable and relatively robust solution can be readily calculated, but at the cost of large numerical diffusion.  Indeed the ability to solve these types of irregular problems helps explain the preference practitioners often demonstrate in choosing low order methods to compute solutions to irregular systems, where to resolve solution features one relies solely on mesh/grid refinement.

In contrast to lower order approaches, the DNN solution to the Sod shock tube problem is able to exploit the relative flexibility of DNN functional approximators, as determined by composition of affine transformations and element-wise activation nonlinearities.   For example, with hyperbolic tangent activations, the DNN solution to the shock tube problem becomes an approximation to a regular solution of the Sod problem at high order accuracy, in the sense that the approximation order of a linear combination of transcendental functions is $C^{\infty}$.

In the case of the DNN solution, even though the discontinuous shock fronts display unbounded derivatives in the classical sense, the DNN is able to find a relatively well-behaved solution as shown in Fig.~\ref{fig:1}.  The DNN formulation with the hyperbolic tangent activation, if the optimization is to converge to a parameter point $\hat{\boldsymbol{\vartheta}}$, is predicated on a regular solution.  Failure to converge could arise if the derivatives at the locus of discontinuity, and the components of $\boldsymbol{\vartheta}$, enter unstable unbounded growth in the iterations of the optimization protocol.  Indeed, the DNN approximates the discontinuous mathematical shock precisely in the limit in which certain components of $\boldsymbol{\vartheta}$ tend to infinity.  We observe that in practice, however, this does not happen.
The inherent spectral bias \cite{1806.08734} of the DNNs seems to be able to contain the potential instability of the solution until the optimization has settled in a ``reasonable'' local minimum.
This leads to an interesting feature of the DNN solutions, namely that they seem to demonstrate robustness while broadly maintaining many of the small-scale features of the exact solution, even in the presence of irregularity in the formulation.


\subsubsection{DNNs as gridless adaptive inverse PDE solvers}\label{Sec4.1.1}

To explain why an approximate regular DNN solution to a shock problem can perform as well as it does, as shown in Fig.~\ref{fig:1}, we note that a DNN solver can really be viewed as a type of gridless adaptive inverse solver.  First, the absence of a fixed grid in the evaluation of the loss functions (\ref{loss}) and determination of the optimization direction (\ref{sgd}) means that the probability density $\wp$ can be finitely sampled in space as densely as the method can computationally afford at every iteration and for as many iterations as can be afforded.  Given the computational resources, this can lead to a relatively dense, adaptively sampled interrogation of the space-time domain.  Moreover, adaptive mesh methods for shock problems have long been known to be highly effective \cite{ ISI:A1989AC01200004,ISI:A1997WL60900007}, particularly when shock fronts can be precisely tracked.  However, AMR and $hp$-adaptive methods are constrained by their meshing/griding geometries, even as isoparametric and isogeometric methods have been developed to, at least in part, reduce these dependencies \cite{MICHOSKI2016658}.

The second pertinent observation about the DNN solution is that because it utilizes an optimization method to arrive at the solution, it can be viewed as a type of inverse problem.  Unlike a forward numerical method, which accumulates temporal and spatial numerical error as it runs, the DNN solutions become more accurate with iterations, being a global-in-space-time inverse-like solver.   This means that it becomes entirely reasonable to consider running such systems in single or even half precision floating point formats, leading to potential performance benefits in comparison to more standard PDE solvers.

\subsection{Diffusing along unbounded gradients}
\label{Sec4.2}

As discussed above, solving a shock problem without adding some form of diffusion (often in the form of limiters or low order accurate methods) tends to lead to unstable solutions in classical numerical methods.  Similarly here, even in the case of the DNN solution, when searching for an approximation to the regular solution shown in Fig.~\ref{fig:1}, the resulting solution, though reasonably well-behaved, is of course spurious along the undefined shock fronts.   

As in all numerical methods, to ameliorate this problem, all that needs to be done is to add some small amount of diffusion to constrain the derivatives along the shock fronts.  In this case, we slightly alter (\ref{euler}) into the non-conductive compressible Navier-Stokes type system, \begin{equation}\begin{aligned} \label{NS} & \partial_{t}\boldsymbol{U} + \partial_x \boldsymbol{F}(\boldsymbol{U}) - \partial_{x}\boldsymbol{G}(\boldsymbol{U}) = 0, \quad \boldsymbol{U}_{|t=0} = \boldsymbol{U}_{0}, \quad \boldsymbol{U}_{|x\in\partial\Omega} = \boldsymbol{V}_{\partial\Omega}, \end{aligned}\end{equation} given a dissipative flux $\boldsymbol{G}(\boldsymbol{U}) = (0, \rho \tau u_x, \tfrac{1}{2} \rho \tau u^2_x)^{\top}$.  Again here we start by merely solving over the space-time domain, $\Omega \times [0,T_{s}]$, given a fixed and positive constant $\tau\in\mathbb{R}^{+}$.  As we show in more detail below, adding a modest amount of viscosity $\tau=0.005$ leads to a DNN solution that is competitive with some of the most effective numerical methods available for solving the Sod shock tube problem.

\section{Numerical Comparison Tests}\label{Sec5}

In this section we show numerical comparison tests between DNNs, the DG finite element methods (FEMs), and FVMs for the Sod shock tube problem.  We show solutions in the eyeball norm for each, relative to, first, grid density, and then degrees of freedom in the method.  It should be noted that accuracy in the DNN solution is somewhat arbitrary, as they can be  tuned via metaheuristics to search for local minima that improve upon basic accuracy measures, such as $L^p$ error.  This is in contrast to, for example, DG and FVM methods which have established convergence criteria, so that as a function of spatial order and mesh width expected improvements in smooth solutions can be anticipated.

\begin{table}[t]
  \caption{Mean square error of the unlimited DNN solution in Fig.~\ref{fig:1} and the solutions as a function of the grid density in Fig.~\ref{fig:2}, against the exact Sod shock tube solution.}
  \begin{tabular}{|c|c|c|c|c|c|c|c|c|}
    \hline
     Solution Type & Density & Pressure & Velocity   \\
    \hline\hline
     Unlimited DNN & 0.00199 & 0.00074 & 0.01272 \\
    \hline
    FVM 1st Order & 0.00117 & 0.00139 & 0.00804 \\
    \hline\hline
    FVM 2nd Order & 0.00030 & 0.00019 & 0.00164 \\
    \hline
    DG, $p=2$ & 0.00163 & 0.00259 &  0.01413 \\
    \hline
    DG, $p=5$  & 0.00314 & 0.00811  & 0.02945  \\
    \hline
    DNN, $\tau=.005$ & 0.00020 & 0.00025 & 0.00136 \\
    \hline

\end{tabular}
    \label{table:2}
\end{table}

The DNN solution method is effectively adaptive and gridless and relies on stochastic optimization to establish minimizing solutions.  In the standard DNN formulation of (\ref{euler}) a list of tunable parameters would include a choice of activation function $\phi$, the number of optimization iterations $l$, the number and distribution of sampled points from $\wp$, the optimization method stopping criteria, the form and weighting of the loss functions in $J_i$, the learning rate of the optimizer $\eta$, etc.  The result is that comparing the accuracy of a DNN solution to a more standard forward solver in some specified norm is not particularly insightful (though we still show the MSE of the solutions in Table \ref{table:2} and \ref{table:3}).  But more so, we try to establish some criteria that do not draw us into the quicksand of hyperparameter optimization, by which to compare DNN solutions to FVM and DGFEM solutions below.

\begin{figure}
\includegraphics[width=0.5\textwidth]{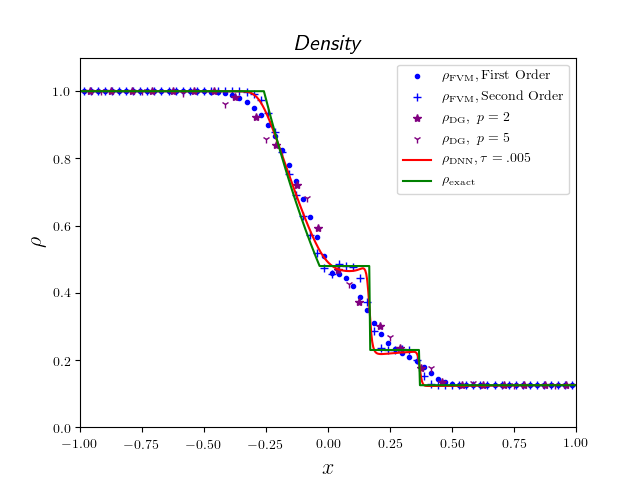}\includegraphics[width=0.5\textwidth]{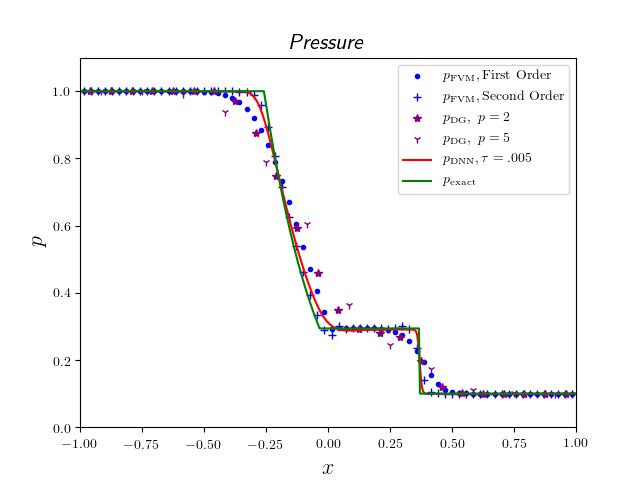} \\ \includegraphics[width=0.5\textwidth]{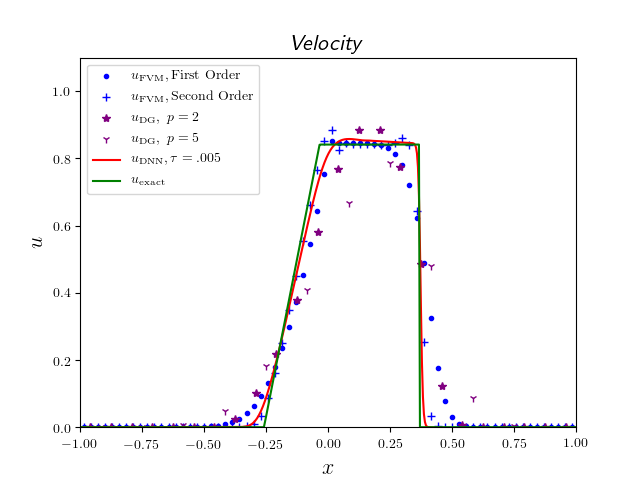}\caption{Comparison of slope-limited discontinuous Galerkin (DG) finite element method solvers, flux-limited finite volume method (FVM) solvers, and the dissipative DNN solver holding fixed the relative resolution in the spatial grids of each solution, $\sqrt{N_{\mathrm{batch}}}\sim N_{\mathrm{vol}}\sim (n+1)N_{\mathrm{el}}$.}\label{fig:2}
\end{figure}

\subsection{Solutions as a function of spatial grid density}\label{gridsec}

One way to evaluate the relationship between the DNN method and FVM or DGFEM is by comparing the relative density of effective residual evaluation points in space.  Broadly, in the DG solution, for degree $n=1$ polynomial solutions, this corresponds to $(n+1)N_{\mathrm{el}}$ for $N_{\mathrm{el}}$ the number of elements.  In the FVM method this corresponds to simply $N_{\mathrm{vol}}$, the number of finite volumes.  In the DNN the residual is evaluated at batch points determined by the sampling distribution $\boldsymbol{x}\sim\wp$, which we denote with $N_{\mathrm{batch}}$. 
One nuance in comparing the methods through residual evaluation points is that in standard FVM and DGFEM methods, these points are spatially fixed.  In the DNN approach however, the batch points are resampled after every  evaluation of the objective function gradients.  As such, it might seem more appropriate to look at how DNNs compare to $hp$-adaptive DGFEM and FVM that use adaptive mesh refinement (AMR). Note, however, that adaptive forward problems using DGFEM and FVM are adaptive in a fundamentally different way relative to the residual evaluation as compared to the DNN method.  In forward problems, the residual evaluation shifts in space relative to the current solution in time.  In the DNN solution, however, though the batch points are resampled, the samples can be chosen independent of the solution behavior, and are done so over the whole spacetime domain $\Omega_h\times [0,T_{s}]$.

\begin{table}[ht]
  \caption{Mean square errors of the solutions as a function of fixed degrees of freedom in both space and time, as shown in Fig.~\ref{fig:3}.}
  \begin{tabular}{|c|c|c|c|c|c|c|c|c|}
    \hline
     Solution Type & Density & Pressure & Velocity   \\
    \hline\hline
    FVM 1st Order & 0.00027 & 0.00017 & 0.00136 \\
    \hline\hline
    FVM 2nd Order & 0.00011 & 0.00005 & 0.00084 \\
    \hline
    DG, $p=2$ & 0.00021 & 0.00013 & 0.00182  \\
    \hline
    DG, $p=5$  & 0.00042 & 0.00029 & 0.00257  \\
    \hline
    DNN, $\tau=.005$ & 0.00020 & 0.00025 & 0.00136 \\
    \hline

\end{tabular}
    \label{table:3}
\end{table}

This dependence on space and time of the sampled batch points $N_{\mathrm{batch}}$, distinguishes the DNN solution from even the adaptive FVM and FEM solutions.  Thus in order to compare the spatial resolution of the two different types of solutions, we compare solutions using the rule of thumb that $\sqrt{N_{\mathrm{batch}}}\sim N_{\mathrm{vol}}\sim (n+1)N_{\mathrm{el}}$.

Results of this comparison are shown in Fig.~\ref{fig:2} and Table \ref{table:2}, where the timestepping is chosen sufficiently small.  Here the DGFEM solution is run at $n=2$, $N_{\mathrm{el}}=24$ and at $n=5$, $N_{\mathrm{el}}=12$ using a Rusanov Riemann solver with adaptive hierarchical slope limiting \cite{ ISI:000302933100009}.  Both first and second order FVM solutions are run at $N_{\mathrm{vol}}=72$, where the first order solution is unlimited, and the second order uses a standard per-face slope limiter.  The DNN solution is run at $\sqrt{N_{\mathrm{batch}}}=72$, using $\tau=0.005$.  The results indicate that the DNN solution performs favorably to the FVM and DGFEM solutions relative to spatial grid density.

\begin{figure}
\includegraphics[width=0.5\textwidth]{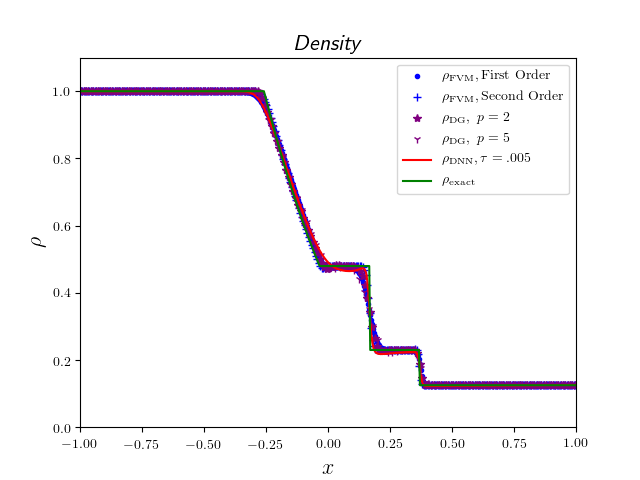}\includegraphics[width=0.5\textwidth]{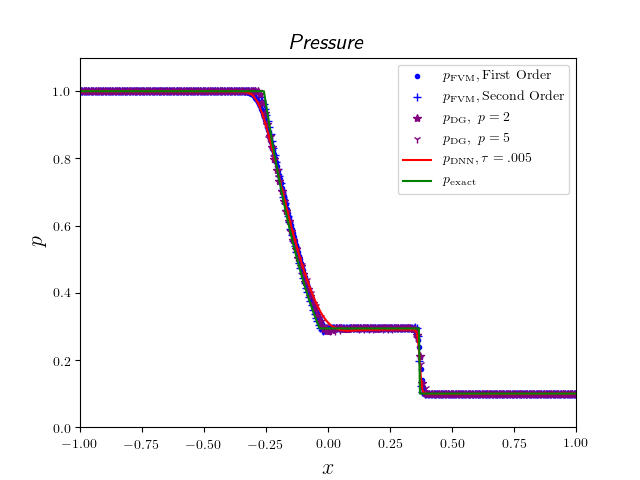} \\ \includegraphics[width=0.5\textwidth]{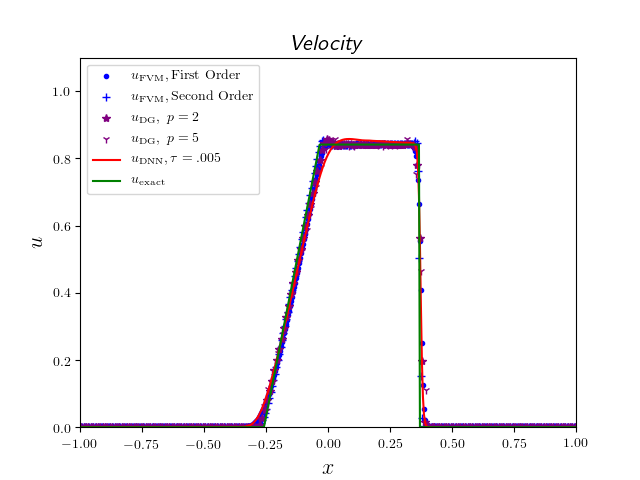}\caption{Comparison of slope-limited discontinuous Galerkin (DG) finite element method solvers, flux-limited finite volume method (FVM) solvers, and the dissipative DNN solver holding fixed the total degrees of freedom (in both space and time) of the discrete representation.  }\label{fig:3}
\end{figure}

\subsection{Solutions as a function of degrees of freedom}

Another way of comparing the DNN solution to the FVM and DGFEM methods is to look at solutions as a function of their respective degrees of freedom (DoFs).  Heuristically this can be thought of as comparing the number of ``tunable parameters'' within the representation space of the solution.

Since the solution in the DNN is a global in space and time solution, we consider the DoFs spanning the whole space, $\Omega_h\times [0,T_s]$.  In the DG solution this corresponds to $S(n+1)N_{\mathrm{el}}T_{\mathrm{grid}}$, where $T_{\mathrm{grid}}$ is the temporal grid and $S$ are the number of stages in the Runge-Kutta discretization.  Likewise in the FVM setting, we have  $SN_{\mathrm{el}}T_{\mathrm{grid}}$.  In the case of the DNN, on the other hand, the degrees of freedom are recovered by counting the scalar components of the vector $\boldsymbol{\vartheta}$ parametrizing the network architecture (\ref{DNNs}), and this yields \begin{equation}\label{DNNdofs}\mathrm{DNN}_{\mathrm{DoFs}} = N_{1}(d_{\mathrm{in}}+1)+d_{\mathrm{out}}(N_L +1) + \sum_{\ell=1}^{L-1} N_{\ell+1}(N_{\ell}+1),\end{equation} where $L$ is the number of hidden layers, $N_{\ell}$ is the width of the $\ell$-th layer, and $d_{\mathrm{in}}$ and $d_{\mathrm{out}}$ are the input and output dimensions, respectively.

The results are presented in Fig.~\ref{fig:3} and Table \ref{table:3}.  In the DNN solution, the degrees of freedom are $\mathrm{DNN}_{\mathrm{DoFs}}=248065$ after setting $L=16$ and $N_{\ell}=128$.  The DGFEM solutions are run the same as in section \ref{gridsec}, except that at sixth order, $n=5$, $S=4$, $N_{\mathrm{el}}=104$ and $T_{\mathrm{grid}}=100$, so that $\mathrm{DGFEM}^{(6)}_\mathrm{DoFs}=249600$.  Similarly the third order accurate DGFEM solutions use $n=2$, $S=4$, $N_{\mathrm{el}}=208$ and $T_{\mathrm{grid}}=100$, corresponding to $\mathrm{DGFEM}^{(3)}_\mathrm{DoFs}=249600$.  The first order accurate FVM solution is run the same as in section \ref{gridsec}, except $S=1$ (a forward Euler solver is used), $N_{\mathrm{vol}}=500$, and $T_{\mathrm{grid}}=500$ so that $\mathrm{FVM}^{(1)}_\mathrm{DoFs}=250000$.  At second order the FVM method uses a predictor-corrector method, which we set as $S=2$, $N_{\mathrm{vol}}=350$, and $T_{\mathrm{grid}}=355$ corresponding to $\mathrm{FVM}^{(2)}_\mathrm{DoFs}=248500$.  Again, the results show that the DNN solution compares favorably as a function of degrees of freedom.

\section{Natural Extensions}\label{Sec6}

In this section we discuss a pair of natural extensions to the DNN framework applied to the Sod Shock tube problem.  The first of these is an application of a type of ``simulated annealing," that can be used along the viscous and/or dissipative parameter directions to improve the effective time to convergence in the method.  Next we show how even more powerfully, solving the DNN framework along a full parameter axis, in this case along the viscous axis $\tau$, can lead to a simultaneous and fully parameterized solution in the entire parameter space at nearly no additional cost.

\subsection{Dissipative annealing}\label{Sec:6a}
\begin{figure}
\includegraphics[width=0.5\textwidth]{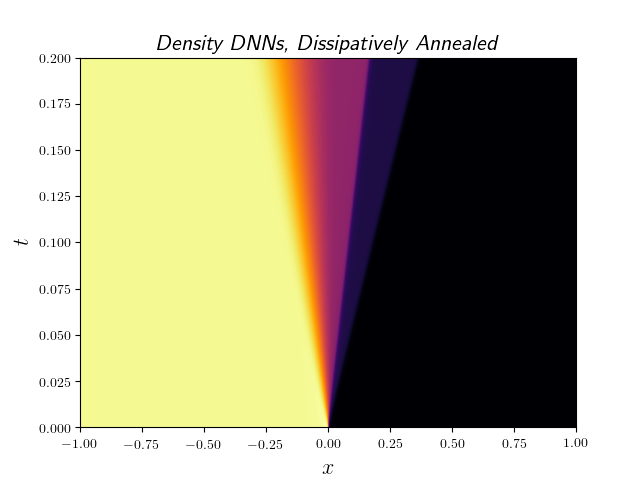}\includegraphics[width=0.5\textwidth]{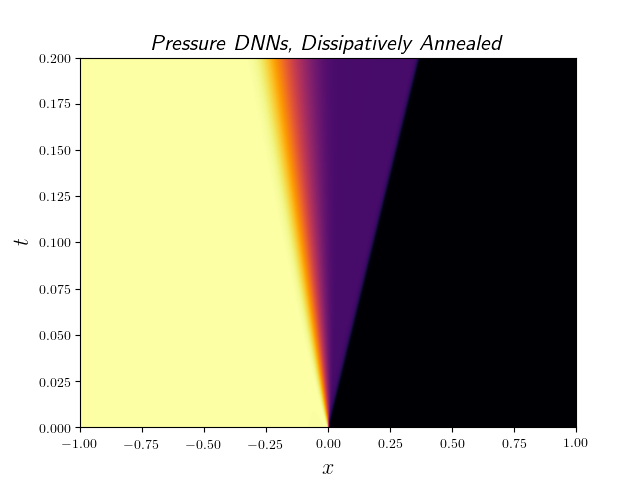} \\ \includegraphics[width=0.5\textwidth]{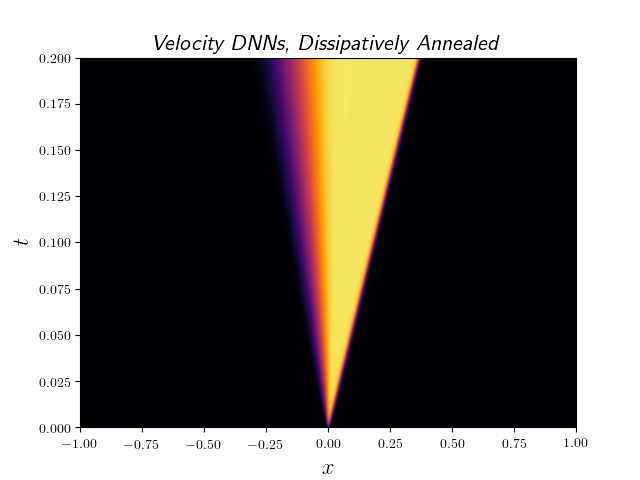}\caption{The Sod shock tube solution using a DNN solver along with dissipative annealing, out to only $60$K iterations.}\label{fig:4}
\end{figure}

\begin{figure}
\includegraphics[width=0.5\textwidth]{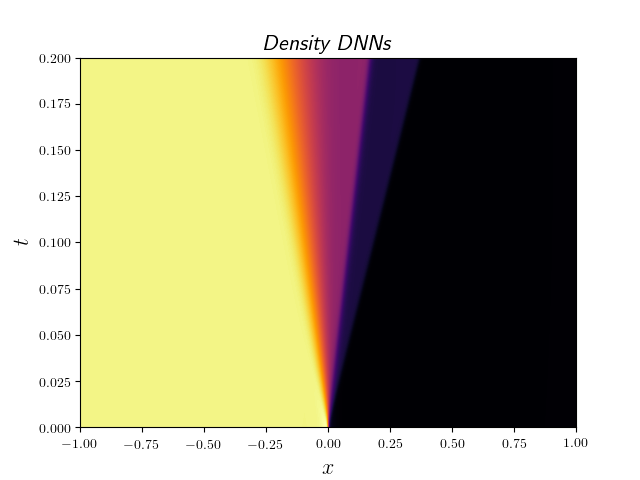}\includegraphics[width=0.5\textwidth]{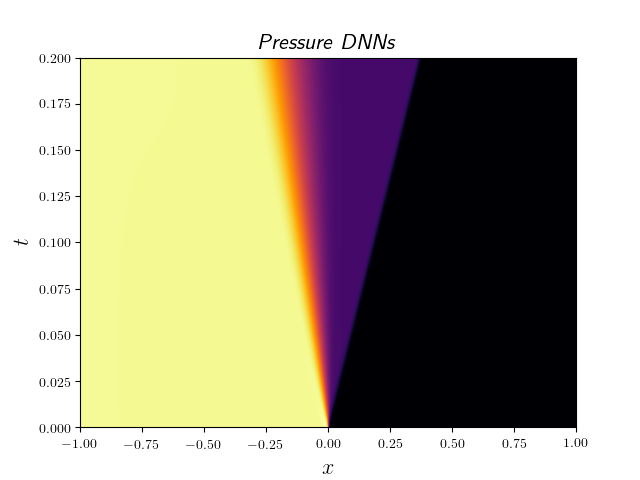} \\ \includegraphics[width=0.5\textwidth]{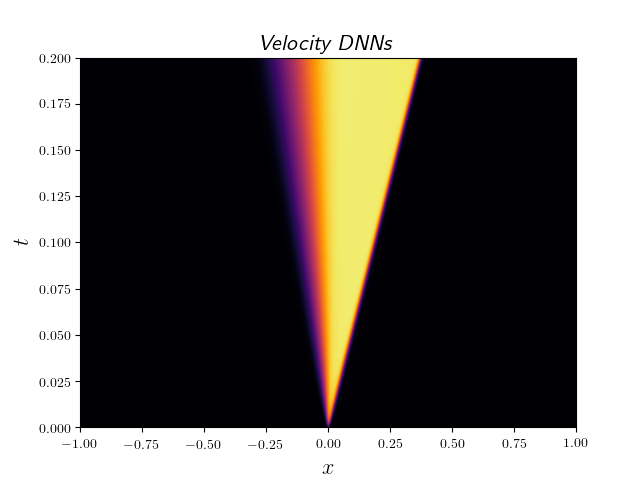}\caption{The Sod shock tube solution using a DNN solver \emph{without} dissipative annealing to $100$K iterations.}\label{fig:5}
\end{figure}

\begin{figure}
\includegraphics[width=0.5\textwidth]{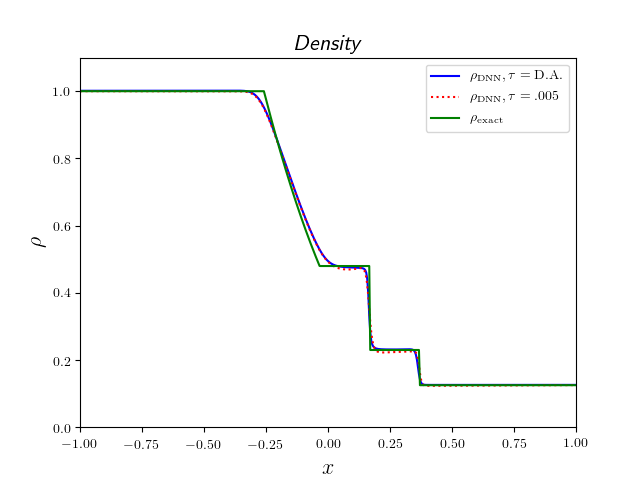}\includegraphics[width=0.5\textwidth]{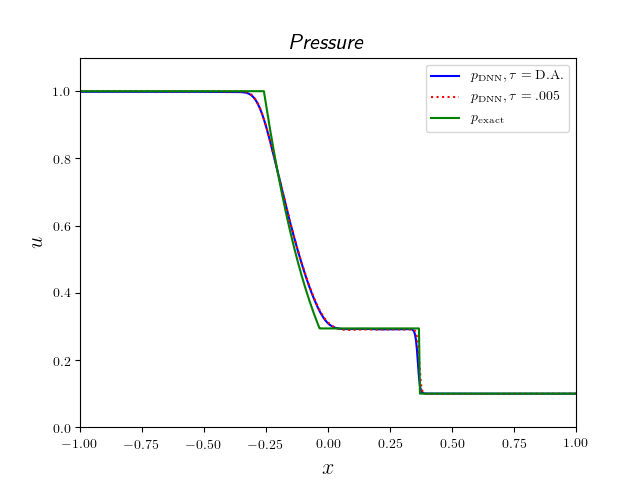} \\ \includegraphics[width=0.5\textwidth]{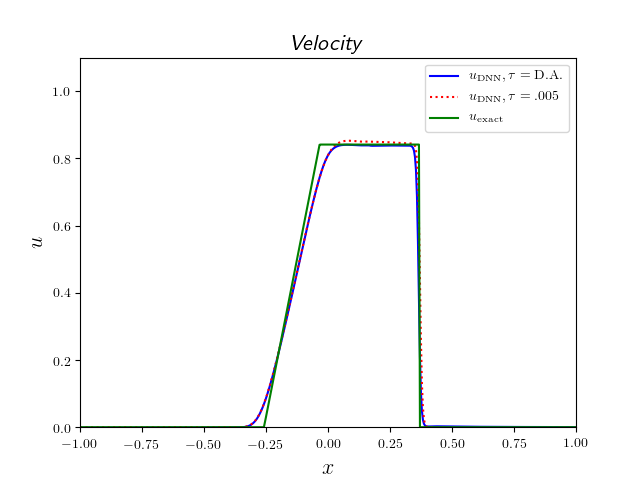}\caption{The final timestep of Sod shock tube solution comparing the dissipative annealed solution at $60$K, to the standard solution at $100$K iterations, to the exact solution.}\label{fig:6}
\end{figure}

One of the potential limitations of the DNN approach, in comparison to other numerical methods, is the compute time-to-convergence.  Discussion of convergence in the DNN/PDE setting is complicated by the fact that ``convergence'' refers to different concepts in numerical PDE analysis and in numerical optimization.  In numerical PDE analysis, convergence usually refers to rate at which an approximate solution tends to the exact solution as a function of the representation parameters, e.g., mesh width $h$ and/or polyniomial order $p$. In iterative numerical optimization, however, convergence refers to the rate at which the parameters $\boldsymbol{\vartheta}$ converge to a (possibly local) minimum as a function of the number of iterations.  The global minimum may exist only in the limit in which some components of $\boldsymbol{\vartheta}$ tend to infinity. Then, heuristics mandates that we seek convergence to an acceptable local minimum.  It is thus difficult to arrive at a formal definition for what is meant by time-to-convergence in the DNN setting.  In this section, we resort to a hybrid heuristic between the two concepts of convergence discussed above, and refer to ``time-to-convergence'' as the number of computational cycles required to reach a comparable level of accuracy to that of a more traditional PDE solution method.  It is in this sense that DNN solutions can appear noticeably more expensive, though in this section we discuss some of the nuances that underlie this observation, and why the unique capacity and flexibility of DNN solutions makes the time-to-convergence less restrictive than it seems.

\begin{figure}
\includegraphics[width=0.49\textwidth]{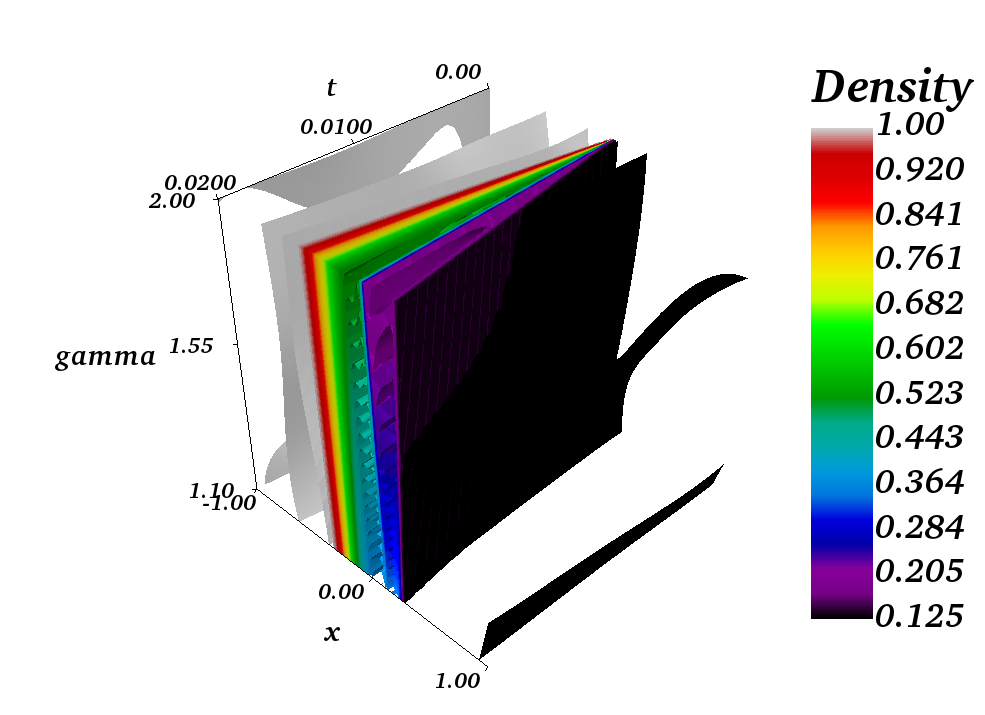}\includegraphics[width=0.49\textwidth]{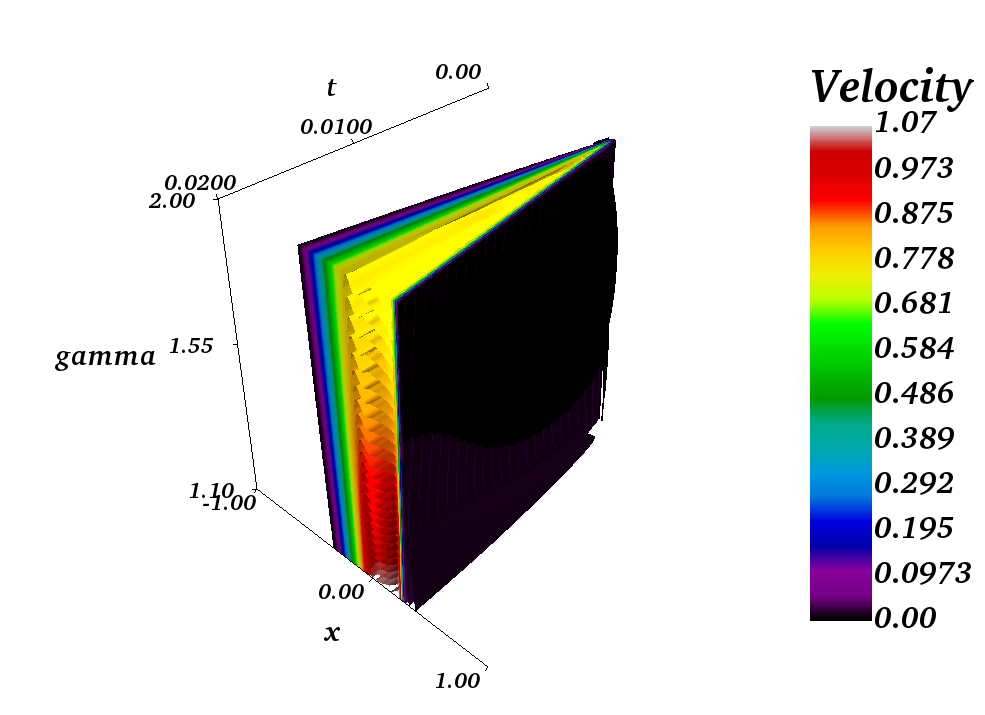} \\ \includegraphics[width=0.49\textwidth]{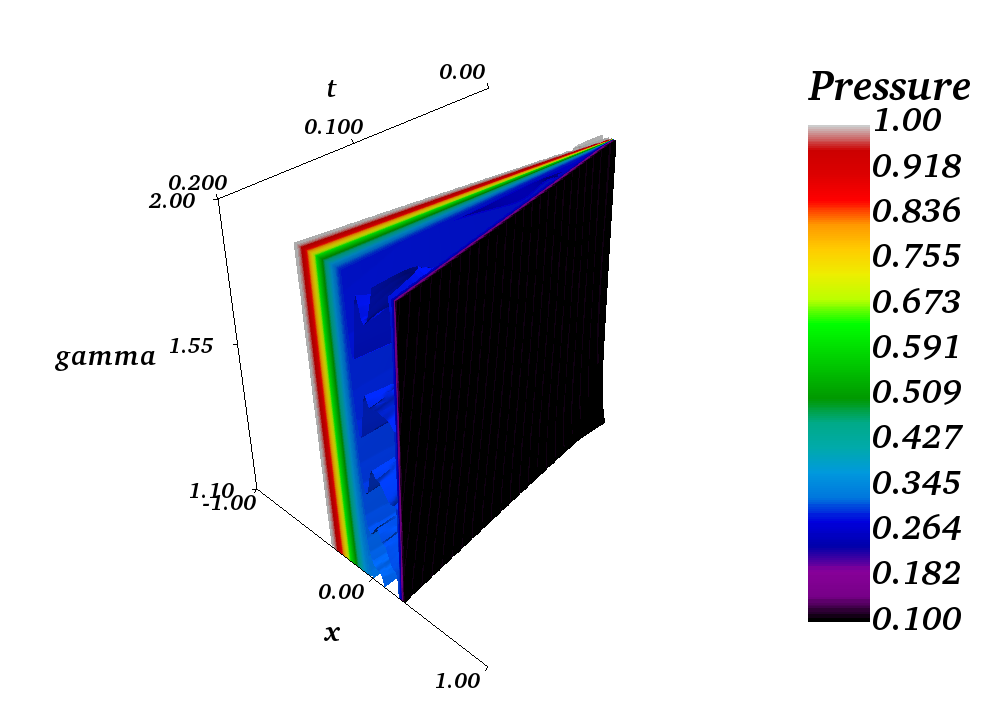} \includegraphics[width=0.49\textwidth]{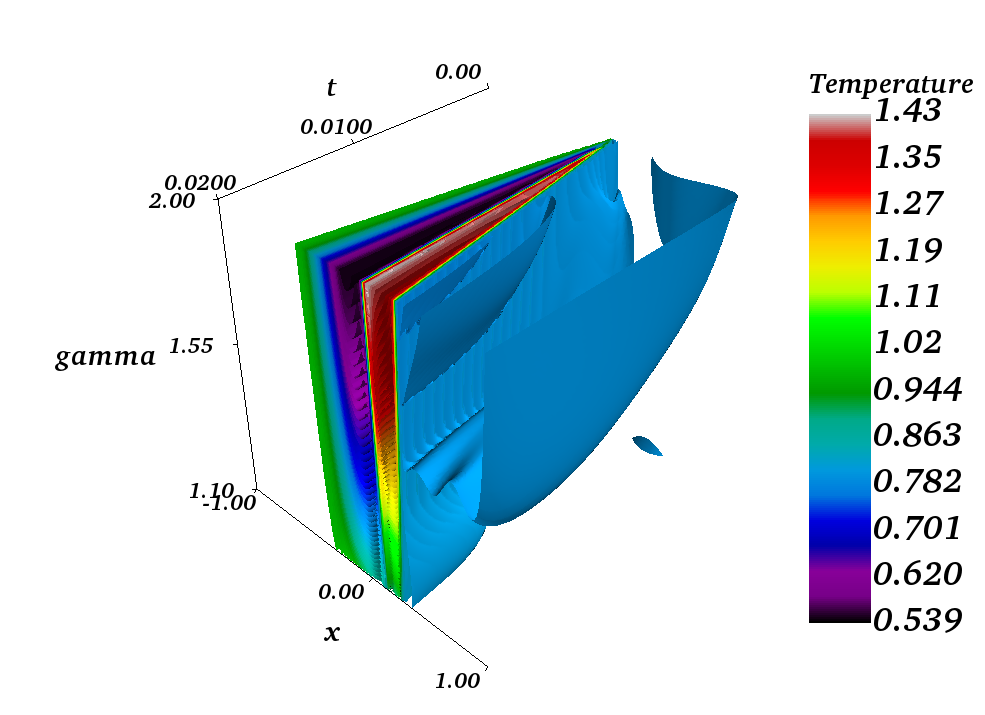}\caption{100 3D contours of the Sod shock tube solution solved over the continuous parameter scan, $(x,t,\gamma)\in [-1,1]\times[0,0.2]\times[1.1,2.0]$ at 100K iterations.} \label{fig:3b}
\end{figure}

\begin{figure}
\includegraphics[width=0.49\textwidth]{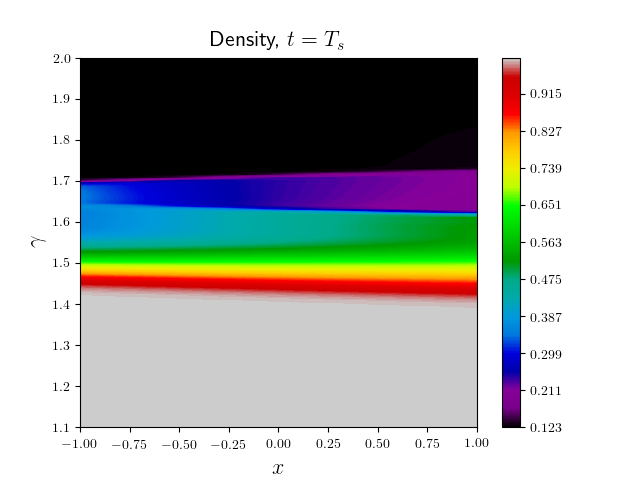}\includegraphics[width=0.49\textwidth]{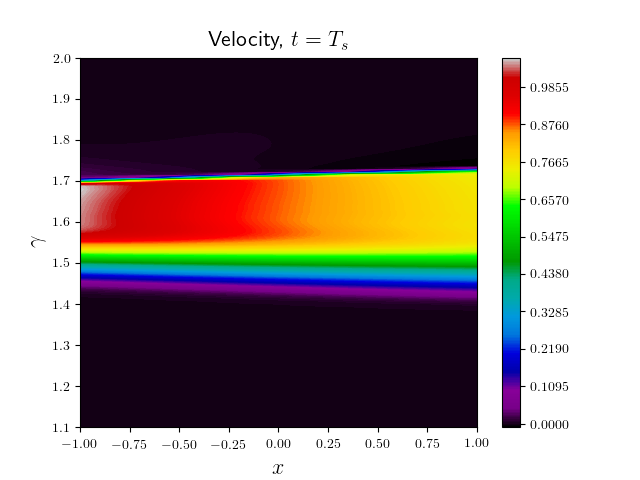} \\ \includegraphics[width=0.49\textwidth]{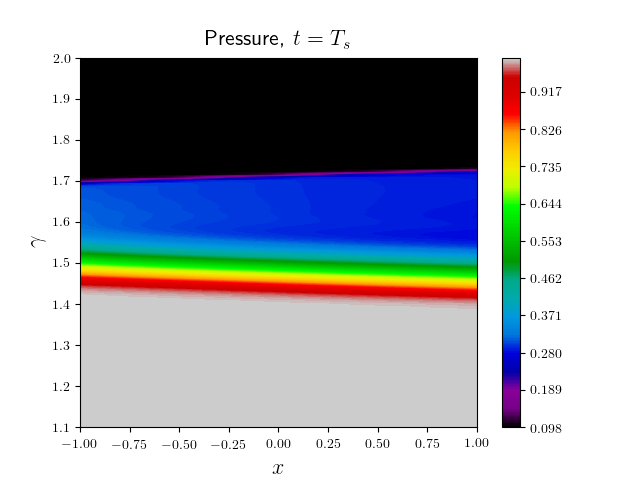} \includegraphics[width=0.49\textwidth]{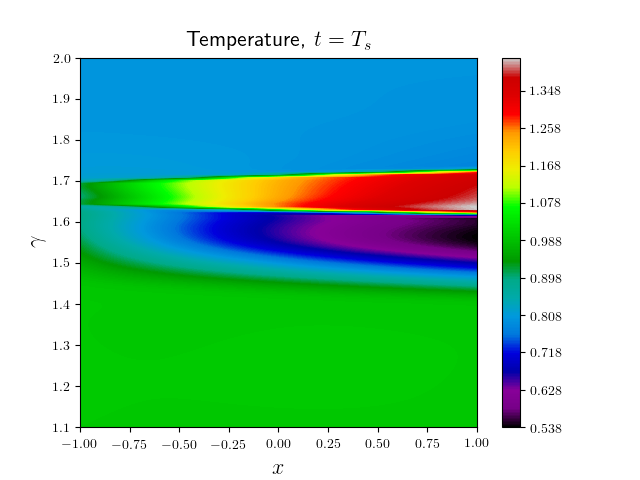}\caption{Variation in $\gamma$ and $x$ of the Sod shock tube solution, at $T_s=0.2$ with 100K iterations.} \label{fig:3c}
\end{figure}

The premise we assume here about slowly converging solutions (in the sense of optimization) is that large-scale features of a smooth solution are relatively easy to converge to, but when accompanied with finer scale features, the optimization noise from attempting to fit the finer scales can obscure the larger scale features from the optimizer.  To mitigate this effect,  inspired by the simulated annealing method \cite{Kirkpatrick1983OptimizationBS}, we recast the dissipative flux in (\ref{NS}) as \[\boldsymbol{G}(\boldsymbol{U}) = (0, \rho \tau_{0} u_x, \tfrac{1}{2} \rho \tau_{0} u^2_x)^{\top},\] where $\tau_{0}=\tau_{0}(l)$ becomes an iteration-$l$ numerical viscosity.  In this case $\tau_{0}(l) = g(l)\tau_{\mathrm{smooth}}$ for some smooth $\tau_{\mathrm{smooth}}\in\mathbb{R}$.  The function $g(l)$ is chosen as a fractional stepwise function bounded from above by unity.

The idea of using $\tau_0$, is to converge early and fast in the iteration cycle to an overtly smooth solution of (\ref{NS}) with large viscosity.  Since such a solution has dramatically fewer fine-scale features, it is conjectured that the optimizer can more easily find a stable minimum of the objective function for such a smooth solution.  As a consequence, fewer total iterations are required to converge to the minimum.

We have tested this idea on the Sod problem, and have found that with minimal effort it seems to reduce the number of iterations needed to arrive at similar results.  For example, in Figs.~\ref{fig:4}--\ref{fig:6}, we test a DNN using a decreasing learning rate schedule.  This base case is run with the minibatch size of $5000$ and initial learning rate of $10^{-4}$ that is reduced every 25000 iterations by factor of $0.1$ for a total of 100000 iterations.  This case is compared to a dissipatively annealed solution obtained with the same minibatch size, but instead of $25000$ iterations per learning rate decrement we find we can get away with $12000$ iterations, where after each boundary set, $g=\{1,0.2,0.9,0.77,0.714\}$, such that $\tau_0(l) = g(l)\tau_{\mathrm{smooth}}$ for $\tau_{\mathrm{smooth}}=0.05$ and $l=1,\ldots,5$.  This means the dissipatively annealed solution takes only 60000 iterations to arrive at $\tau_{0}(5) = 0.005$.

As is clear from Figs.~\ref{fig:4}--\ref{fig:6}, the results are largely indistinguishable, where in some ways the dissipatively annealed solution actually looks better for 60\% of the computational cost.  This result, however, should be taken with a grain of salt.  In this example case the two runs are set up exactly the same, up to the dissipative anealing algorithm.  It is not clear that this is an entirely fair comparison, given the number of hyperparameters that can be tuned in each case. 

\subsection{Simultaneous parameter scans}

Dense parameter space exploration is the most reliable way to develop predictive input-output mappings for scenario development, optimal design and control, risk assessment, uncertainty quantification, etc., in many real-world applications.  Without understanding the response to the parameters of the system, be they variable boundary conditions, interior shaping constraints, or constitutive coefficients, it is all but impossible to examine the utility an engineered model might have in some particular design process.  PDEs clearly help in this regard since they are fully parametrized models of anticipated physical responses.  That being said, PDEs are also often fairly expensive to run forward simulations on, frequently requiring large HPC machines to simulate even modest systems \cite{Bremer2019,Malaya:2017:EPS:3093338.3093371}. Consequently, parameter space exploration of a PDE system is usually both very important and very computationally expensive.

To mitigate the expense of running forward models, surrogate modeling \cite{ trove.nla.gov.au/work/7408865} and multifidelity hierarchies \cite{DBLP:journals/siamrev/PeherstorferWG18} are often conceived in order to develop cost-effective ways of examining the response of a PDE relative to its parametric responses.   One of the ways in which these methods function is by developing reduced models to emulate the parametric dependencies from a relatively sparse sampling of the space.  In this way it becomes possible---for a sufficiently smooth response surface---to tractably parametrize the input-output mapping, and thus interrogate the response at a significantly reduced cost, lending itself to statistical inference, uncertainty quantification, and real-world validation studies.

However, as is commonly the case, the response surface is not sufficiently smooth, or its correlation scale-length cannot be determined a priori, and these reduced order mappings can become highly inaccurate.  In such circumstances having a way to emulate the exact response surface would clearly be of high practical value.   As a potential solution to this, one of the most powerful immediate features of the DNN framework seems to be the ability to perform simultaneous and dense parameter space scans with both very little effort. The framework is thus  a natural, and in some sense exact emulator for complicated system response models.  In the case of the Sod shock problem this can be accomplished by simply recasting (\ref{NS}) relative to its parameter $\gamma$, where instead of solving over space-time $(x,t)\in\Omega\times [0,T_s]$, the solution is solved over the parameter-augmented parameter space $(x,t,\gamma)\in \Omega\times [0,T_s]\times[\gamma_{\mathrm{low}},\gamma_{\mathrm{high}}]$.

Intuitively one might expect the increase of dimensionality to be prohibitive, as the parameter space grows from a discrete 2D system to a discrete 3D system.  However, in analogy with the observation of section \ref{Sec:6a} where the parameter $\tau$ is dissipatively annealed, this is not what is observed.  Instead, using the same DNN parameter sets, the same minibatch size and learning rate schedule, optimization over the parameter-augmented input space $(x,t,\gamma)$ reaches similar loss values with almost no computational overhead.  Contour and time slice plots of the solutions are shown in Fig.~\ref{fig:3b} and Fig.~\ref{fig:3c} for the density, velocity pressure, and temperature, each as a function of $(x,t,\gamma)\in [-1,1]\times[0,0.2]\times[1.1,2.0]$. 

Remarkably, in this example, the DNN framework demonstrates that it is as cost effective to perform dense parameter space exploration along the dimension of the adiabatic index $\gamma$ as it is to solve at a single fixed value of $\gamma$.  It remains unclear how robust this behavior is over physical parameters of the model --- in this case (\ref{NS}).  But however robust this behavior ends up being, even if only across isolated and specific parameters, this demonstrates a remarkably advantageous aspect of the DNN formulation.



\section{DNNs With Residual Connections and Multiplicative Gates}\label{sec:LSTM}

As pointed out in \cite{SIRIGNANO20181339}, the DNN architecture can influence its behavior. In \cite{SIRIGNANO20181339} it is claimed that an LSTM-inspired feed-forward network architecture with residual connections and multiplicative gates can help improve performance for some parabolic PDE problems.  These models are effectively an alternative to the more standard deep neural network architecture presented in (\ref{DNNs}).

We implement these LSTM-like architectures for our 1D mixed hyperbolic-parabolic like PDE system (\ref{euler}), to see how it behaves relative to standard DNNs in the context of more irregularity in the solution space.  Setting $\boldsymbol{\zeta} = (\boldsymbol{x},t)$, we test the Euler system (\ref{euler}) on the following architecture comprised of $L+1$ hidden layers: \begin{equation}\begin{aligned}\label{LSTMlike} S^0 & = \phi(W^0\boldsymbol{\zeta} + b^0), \\ Z^\ell & = \sigma ( U^{z,\ell}\boldsymbol{\zeta} + W^{z,\ell}S^\ell+b^{z,\ell}),  \quad \ell = 0,\ldots, L-1, \\ G^\ell & = \sigma (U^{g,\ell}\boldsymbol{\zeta}+W^{g,\ell}S^\ell+b^{g,\ell}), \quad \ell = 0,\ldots, L-1, \\ R^{\ell} & = \phi (U^{r,\ell}\boldsymbol{\zeta} + W^{r,\ell}S^\ell + b^{r,\ell}),   \quad \ell = 0,\ldots, L-1, \\ H^\ell & = \phi (U^{h,\ell}\boldsymbol{\zeta}+W^{h,\ell}(S^\ell\odot R^{\ell}+b^{h,\ell}),   \quad \ell = 0,\ldots, L-1, \\ S^{\ell+1} & = (1-G^{\ell})\odot H^{\ell} + Z^{\ell}\odot S^{\ell},   \quad \ell = 0,\ldots, L-1, \\ z^{L}(\boldsymbol{\zeta},\boldsymbol{\theta}) & = W^LS^{L}+b^L, \end{aligned}\end{equation} where $\sigma(x)=1/(1+e^{-x})$ is the sigmoid function and the network parameters are given by: \begin{equation}\begin{aligned}\label{lstmparams}\boldsymbol{\theta} = \bigg\{& W^0,b^0,\left(U^{z,\ell},W^{z,\ell},b^{z,\ell}\right)^{L-1}_{\ell =0}, \\ & \left(U^{g,\ell},W^{g,\ell},b^{g,\ell}\right)^{L-1}_{\ell =0},\left(U^{r,\ell},W^{r,\ell},b^{r,\ell}\right)^{L-1}_{\ell =0}, \left(U^{h,\ell},W^{h,\ell},b^{h,\ell}\right)^{L-1}_{\ell =0},W^L,b^L \bigg\},\end{aligned}\end{equation} and the number of units per layer is $N_\ell$.  For more details on the network architecture we refer the reader to  \cite{SIRIGNANO20181339}.

\begin{figure}
  \includegraphics[width=0.5\textwidth]{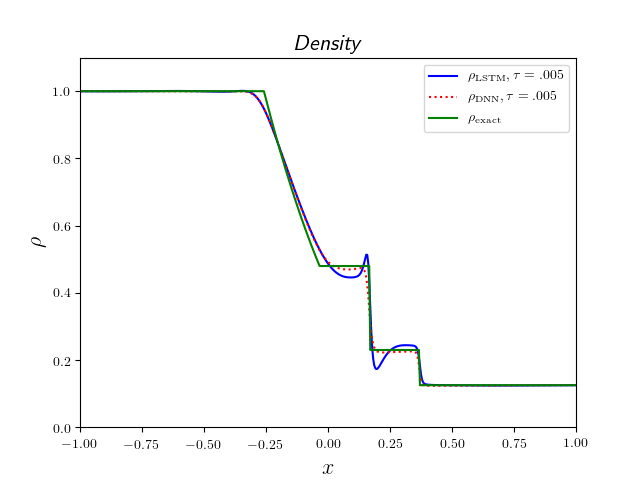}\includegraphics[width=0.5\textwidth]{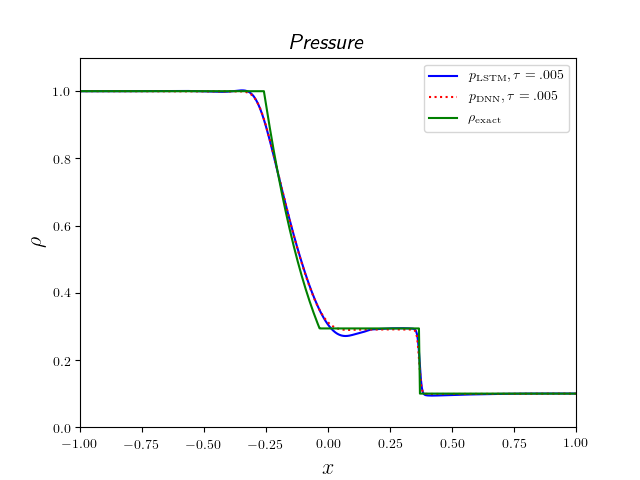} \\ \includegraphics[width=0.5\textwidth]{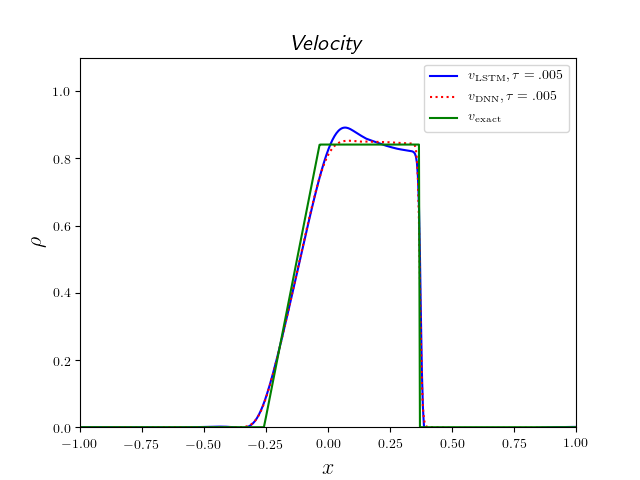}\caption{The Sod shock tube solution at the final timestep, comparing the LSTM-like architecture versus standard DNN at fixed DoFs.}\label{fig:7}
\end{figure}

To understand whether the standard DNN architecture is more effective, in some sense, than (\ref{LSTMlike}), it is essential to recognize that the degrees of freedom scale differently for the LSTM-like system (\ref{LSTMlike}) than they do in (\ref{DNNdofs}), where the LSTM-like system is substantially more expensive per network layer $\ell$.  More explicitly, assuming that each layer has the same width, $N_{\ell}$, the degrees of freedom in the LSTM-like system can be calculated as: \begin{equation}\label{LSTMdofs}  \mathrm{LSTM}_{\mathrm{DoFs}} = 4 L N_{\ell}^2 + (4 L (d_{\mathrm{in}} + 1) + d_{\mathrm{in}} + d_{\mathrm{out}} + 1) N_{\ell} + d_{\mathrm{out}}. \end{equation}

Comparing (\ref{DNNdofs}) to (\ref{LSTMdofs}), for a fixed number of layers $L=2$, setting for (\ref{euler})  $d_{\mathrm{in}}=2$ and $d_{\mathrm{out}}=3$, the resulting relationship between the number of units in the LSTM, $N_{\ell}\implies N_{\mathrm{LSTM}}$, and DNN, $N_{\ell}\implies N_{\mathrm{DNN}}$, can be computed by solving the following ceiling function: \[ N_{\mathrm{DNN}}  =  \ceil[\Bigg]{\frac{-7+\sqrt{16 N_{\mathrm{LSTM}}^2 + 72 N_{\mathrm{LSTM}} + 49}}{2}} .\]

For our numerical test, we set $L=16$ with the reference solution width set to $ N_{\mathrm{DNN}} = 128$.  This leads to $ N_{\mathrm{DNN}} =248451$.  Solving for the relationship above this yields $N_{\mathrm{LSTM}}=63$.  The results are presented in Fig.~\ref{fig:7}, and show fairly unambiguously that for this particular test case the LSTM-like architecture does not perform as well as the standard DNN.  As above, this result must be taken in context.  It may well be, for example, that the tuned parameters for standard DNNs do not tune the LSTM-like networks well, and that the result presented is not a proper comparison.  Again, to fully reveal the relationship between these architectures on even just this one simple hyperbolic test case, would require a full study of its own.

That said, from a purely practical point of view, the implementation of the DNN versus the implementation of the LSTM-like networks seems to lead to a network architecture with a large disparity in the number of graph edges.  The DNN network has effectively one layer per $\ell\in L$, while the LSTM-like network has, in some sense, five layer-like networks per $\ell\in L$ as seen in (\ref{LSTMlike}.  The result of this is that in our implementation, even at fixed degrees of freedom, the LSTM-like network takes almost five times longer in compute time to finish than the standard DNN, at a fixed number of iterations.  Again, it is not clear if this slowdown can be reduced by using a more clever implementation of the LSTM-like networks, or if this is simply a result of increasing the network complexity in the graph.  

\section{Data-enriched PDEs}\label{Sec:enriched}

The simplicity and apparent robustness of DNNs for solving systems of nonlinear and irregular differential equations raises the question: how to incorporate experimental data into such a framework? Work combining DNN-based approach to solving PDEs with conventional supervised machine learning has already started to emerge; we presume that this trend will only accelerate.  Raissi et al. \cite{RAISSI2019686} introduced  ``data discovered PDEs,''  where a relatively traditional parameter estimation is performed over differential operators that are discovered through optimization.   For example, parameter estimation can be performed for $\lambda\in \mathbb{R}$ that factors through a differential operator $\lambda uu_x$. It is easy to see this can be expanded to data driven discovery of PDEs, where parameter hierarchies can be used to generate libraries of operators that are selected based on system specific physical criteria and constraints, in order to effectively ``construct'' and/or discover PDE systems from large data sets  \cite{long2018pdenet}.  These types of approaches are emerging with increasing interest \cite{Rudye1602614,doi:10.1098/rspa.2016.0446,BERG2019239,1904.02107,Raissi:2018:DHP:3291125.3291150}, and it is easy to see how such approaches naturally lend themselves to the DNN frameworks.

These types of empirical PDE discovery techniques offer clear advantages in cases where the systems are assumed to be too complex to easily construct first principles models.  However, researchers also find themselves in a different type of situation, where the physics model that describes the system response is confidently prescribed, so that any mismatch between the model and data raises questions about the experimental setup just as it raises them about the model.  

To explore this common circumstance, we address a situation relating to the experimental validation of a first principles model system.  First principles models are predicated on the idea that the physical dependencies within an experimental system are, or can be, fully understood.  Validation studies, however, frequently discover that this is in fact not the case \cite{ref900446282,OLIVER20151310}.  Frequently systematic behaviors, engineering details, alleatoric and/or epistemic uncertainties can cascade, and lead to systems with behaviors that are uncertain and/or different from those of the model systems anticipated to describe them.

In this circumstance, a researcher might be in a situation where a prescribed model system does not plausibly validate against the large quantities of high resolution experimentally measured data that have been collected.  Some of the scientific questions one might want to raise in such circumstances are: \begin{enumerate}\item Are my experimental results reliable? \item Is my model system sufficient to describe the experimental data? \item How far from my model system is the measured data? \item What is the form of the mismatch between the data and the model? \item How might I enrich my model system to more accurately capture the observed behavior? \item Are there characteristics in the mismatch that reveal missing physical subsystems? \end{enumerate} Below we present a simple example of how to approach such a circumstance, and provide an outline of how such a series of questions can be systematically retired.

\subsection{Hypothetical experimental senario}

We consider the following hypothetical scenario: an experimental test facility is set up to run experiments interrogating the behavior of transonic gas dynamics.  Although experimental setup is mildly exotic, the expected responses of the system are anticipated to obey classical gas dynamics (\ref{euler}). Experiments are run, and a process is initially set up to validate relative to the following dimensionless ideal model system:
 \begin{equation}\begin{aligned}\label{class} \partial_{t}\rho & + (\rho u)_x = 0, \\  \partial_{t}(\rho u) & + \left(\rho u^2 + p -\tau\rho u_x\right)_x = 0, \\ \partial_{t}E & + \left(Eu+pu-\tau\rho u u_x-\kappa T_{x} \right)_x = 0, \end{aligned}\end{equation} where the total energy density is given by \[E = \frac{p}{\gamma -1} + \frac{\rho}{2}u^2,\] and the initial-boundary data is set to the same as in (\ref{euler}).  Here a constant heat conductivity $\kappa=\tau=0.005$ is also assumed.  In the process of validating the system, the classical gas dynamic model (\ref{class}) repeatedly demonstrates that it does not strongly validate against the experimentally measured data.  After making sure the measurement equipment is well calibrated, and the confidence in the data is high, the researchers are left with the quandary: what is happening?

\begin{figure}
  \includegraphics[width=0.7\textwidth]{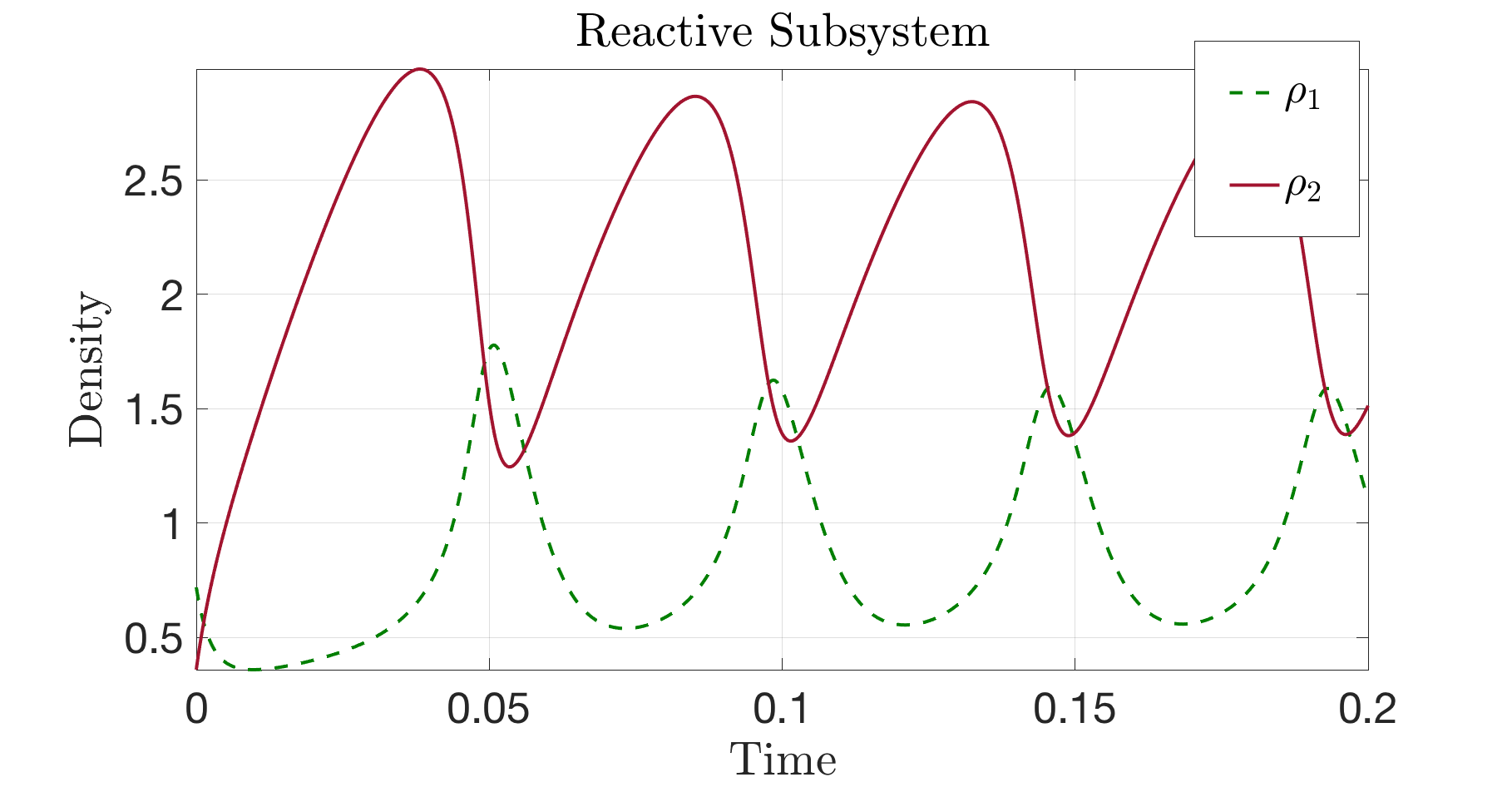}\caption{The reactive subsystem from (\ref{react}).  To illustrate the autocatalytic oscillation, the initial state is set to $\rho_1 = 2 \rho_{l}/3, \rho_{2} = \rho_{l}/3$, though the autocatalysis is largely independent of the initial conditions.}\label{fig:8}
\end{figure}

A traditional approach to this problem might begin with a vigorous debate between the simulation experts and the laboratory experts, both claiming systematic errors on behalf of the other.  
Both camps might proceed by testing their subsystems to the best of their ability and coming to the conclusion that nothing is wrong, per se, but there is some other physics occurring in the test facility that they have not properly anticipated.  As a consequence, the next step might be to slowly start adding back physics terms to (\ref{class}).  Subsequent efforts may include parameter estimation on remaining uncertain parameters in the model until a better fit can be recovered.  In this case, however, such a process would be extremely slow and tedious as illustrated below.

\begin{figure}
\includegraphics[width=0.45\textwidth]{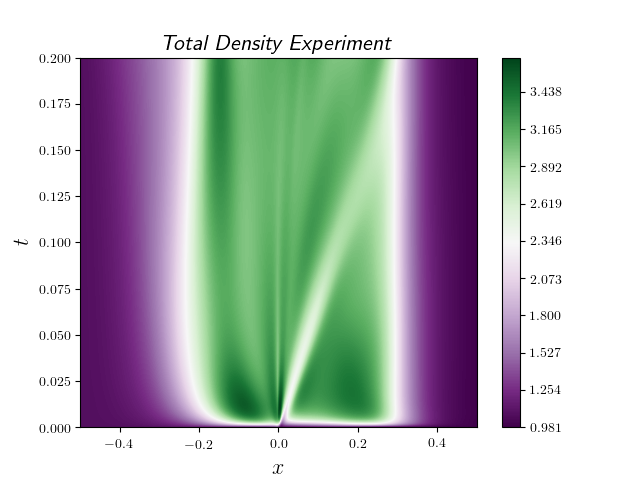}\includegraphics[width=0.45\textwidth]{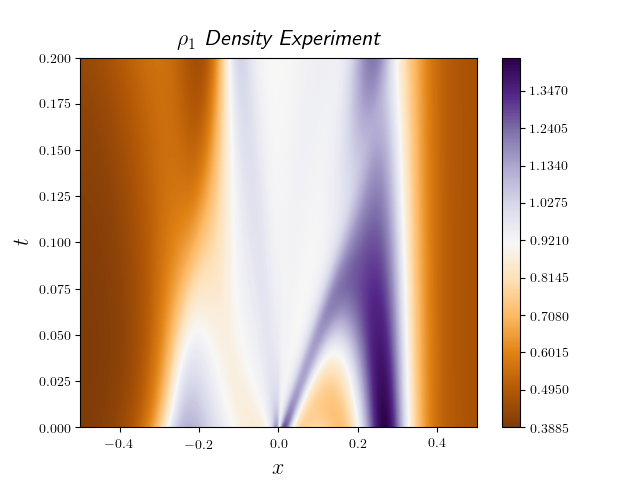} \\ \includegraphics[width=0.45\textwidth]{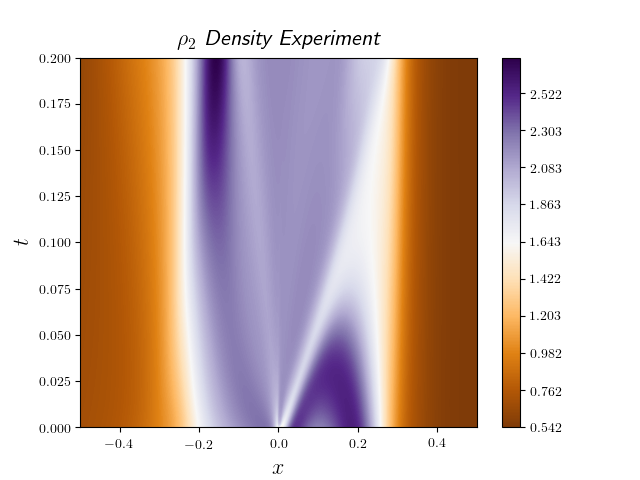} \includegraphics[width=0.45\textwidth]{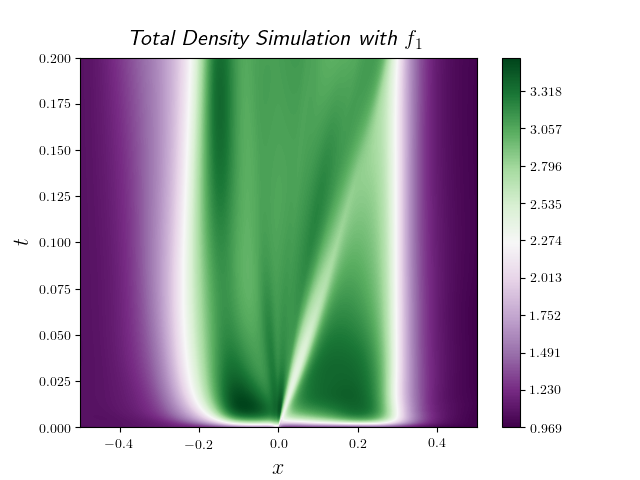}  \\ \includegraphics[width=0.45\textwidth]{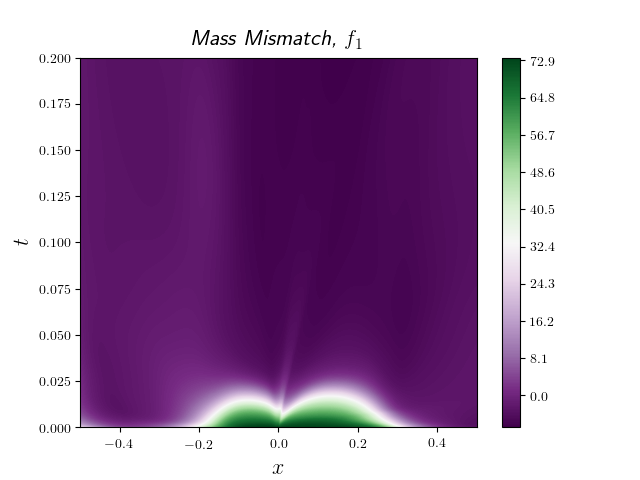} \includegraphics[width=0.45\textwidth]{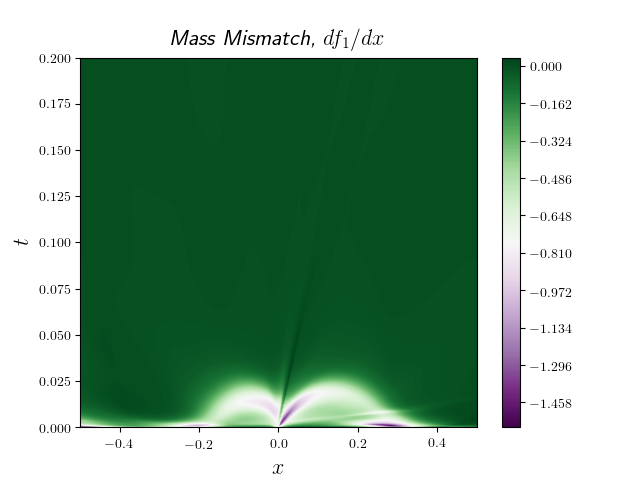}\caption{The top left shows the result from the total measured density in the experiment.  The top right and middle left are the experimentally measured values of the MHD subspecies $\rho_1+\rho_2=\rho$, corresponding to the reactive subsystem in Fig.~\ref{fig:8}.  The middle right is the DNN-enriched simulated value of the density after adding the $f_i$'s.  The bottom panels show the function $f_1$ and it's first derivative, where it's dominant signature effect can be found near the initial state of the system, $t=0$.}\label{fig:9}
\end{figure}

Unbeknownst to both the modellers and experimentalists, the actual dynamics going on inside the experiment, is a substantially more complicated system.  In fact, it turns out that a feature of the experimental setup is inadvertaintly inducing ionization of the gas in the chamber.  As a result, the gas actually behaves like a reactive photoactivated autocatalytic plasma in the center of the reactor, characterized exactly by the following equations: 
\begin{equation}\begin{aligned}
\label{exp} \partial_{t}\rho_1 & + (\rho_1 u)_x = (k_1A_1 + k_2\rho_{1}^{2} \rho_2 - k_{3}\rho_{1}A_2 - k_4\rho_1)\zeta, \\  \partial_{t}\rho_2 & + (\rho_2 u)_x = (k_3\rho_1 A_2 - k_2\rho_{1}^{2} \rho_{2})\zeta, \\  \partial_{t}(\rho u) & + \left(\rho u^2 + p + \frac{1}{8\pi}\boldsymbol{B}^{2}-\frac{1}{4\pi}B_{x}^{2}-\tau\rho u_x\right)_x = 0, \\  \partial_{t}(\rho v)  & + \left(\rho uv  - \frac{1}{4\pi}B_{x}B_{y} - \tau\rho v_x\right)_x = 0, \\  \partial_{t}(\rho w) & + \left(\rho uw  - \frac{1}{4\pi}B_{x}B_{z} -\tau\rho w_x\right)_x = 0,  \\  \partial_{t}E & + \left(Eu+pu+\frac{1}{8\pi}\boldsymbol{B}^{2} - \frac{1}{4\pi}B_{x}(\boldsymbol{v}\cdot\boldsymbol{B})-\tau\rho u u_x-\kappa T_{x} \right)_x = 0, \\ \partial_{t}B_{x}  & = 0, \quad \nabla\cdot\boldsymbol{B}=0, \\ \partial_{t}B_{y} & +\left(uB_{y} - vB_{x}\right)_{x} = 0, \\   \partial_{t}B_{z} & +\left(uB_{z} - wB_{x}\right)_{x} = 0, 
\end{aligned}
\end{equation}
where \[E = \frac{p}{\gamma -1} + \frac{\rho}{2}\boldsymbol{v}^2 +\frac{1}{8\pi}\boldsymbol{B}^2,\quad\mathrm{and}\quad \sum_{i}\rho_i = \rho.\]  Here $\boldsymbol{B} = (B_x,B_y,B_z)$ is a magnetic field, thought inconsequential in the erroneously presumed absence of ions, and the velocity field in the gas is defined as $\boldsymbol{v} = (u,v,w)$.  In this hypothetical scenario, just like for the model system (\ref{class}), the initial conditions for all unknowns are initialized with a jump discontinuity at $x=0$, over the domain $\Omega = [-0.5,0.5]$, for $t\in [0,0.1]$, as inspired in the RP2 case in \cite{doi:10.1002/fld.4681}.  The remaining full boundary conditions are listed in Table~\ref{table:1}.

\begin{figure}
  \includegraphics[width=0.45\textwidth]{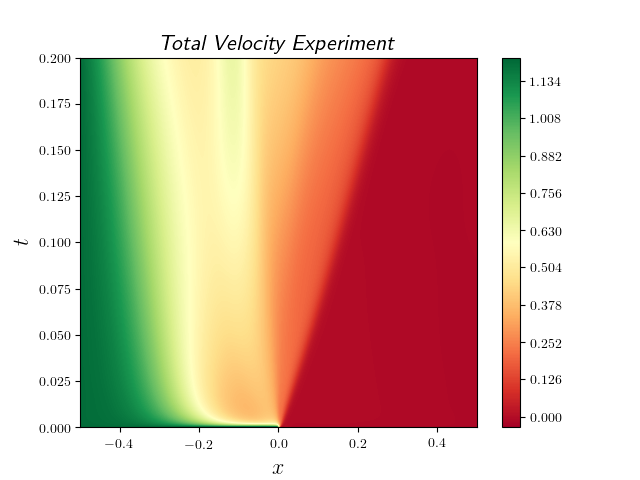} \includegraphics[width=0.45\textwidth]{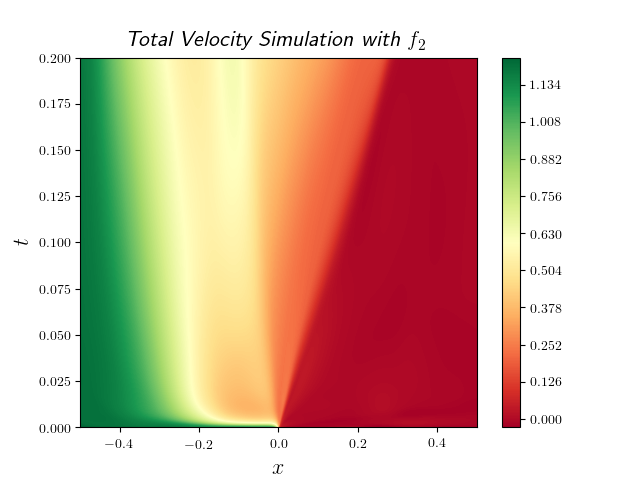}\\  \includegraphics[width=0.45\textwidth]{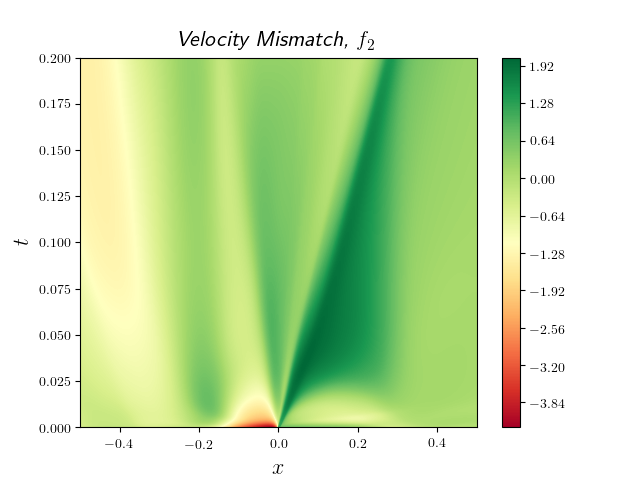} \includegraphics[width=0.45\textwidth]{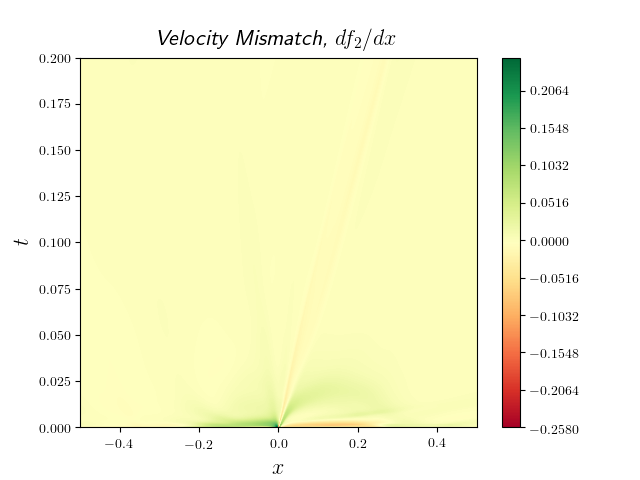} \caption{The top left is the experimentally measured velocity, and the top right the DNN-enriched simulation including the contribution from the mismatch function $f_2$.  The mismatch function $f_2$ is shown on the bottom left, with $df_2/dx$ on the bottom right.}\label{fig:10}
\end{figure}

\begin{table}[t]
  \caption{The initial-boundary data for the model (\ref{class}) and the experiment (\ref{enrich}).}
  \begin{tabular}{|c|c|c|c|c|c|c|c|c|}
    \hline
     B.C. & $\rho$ & $u$ & $v$ & $w$ & $T$ & $B_x$ & $B_y$ & $B_z$  \\
    \hline\hline
     Left Mod. B.C. & 1.08 &  1.2 & -- & --  & 0.8796 & -- & -- & -- \\
    \hline
    Right Mod. B.C. & 0.9891 & -0.0131 & -- & -- & 0.9823 & -- & -- & -- \\
    \hline\hline
    Left Exp. B.C. & 1.08 &  1.2 & 0.01 & 0.5  & 0.8796  & 2.0 & 3.6 & 2.0 \\
    \hline
    Right Exp. B.C. & 0.9891 & -0.0131 & 0.0269 & 0.010037 & 0.9823 & 2.0 & 4.0244 & 2.0026 \\
    \hline

\end{tabular}
    \label{table:1}
\end{table}

The autocatalytic reactive subsystem is induced by virtue of an unexpected ionization pulse in the reactor, leading to an oscillating chaotic attractor characterized by the reactions: \begin{equation}\begin{aligned}\label{react} A_1  & \ce{->T[{$ \ \ \ k_{1} \ \ $}] } \rho_{\mathbbm{1}} \\ 2 \rho_{\mathbbm{1}} +  \rho_{\mathbbm{2}}  & \ce{->T[{$ \ \ \ k_{2} \ \ $}] } 3 \rho_{\mathbbm{1}}  \\ A_2 +  \rho_{\mathbbm{1}} & \ce{->T[{$ \ \ \ k_{3} \ \ $}] } \rho_{\mathbbm{2}} + A_3 \\ \rho_{\mathbbm{1}}  & \ce{->T[{$ \ \ \ k_{4} \ \ $}] } A_4\end{aligned}\end{equation}  where $\rho_{\mathbbm{1}}$ is the first chemical species with density $\rho_1$, and $\rho_{\mathbbm{2}}$ the second chemical species with density $\rho_2$.  Here the $A_i$ are excess bulk species, and the $k_{i}$ are dimensional rate constants.  The condition for instability is that $A_2 > A_1^2+1$, thus we set $A_2 =2$, and $A_1=0.9$, where for simplicity we set $k_i = 150$ for all $i$.  Photoactivation only occurs in the center of the reactor, and thus \[\zeta = \frac{1}{12\sigma\sqrt{2\pi}} e^{-\tfrac{x^2}{2\sigma^2}}.\]  The solution to this subsystem is shown in Fig.~\ref{fig:8}

\begin{figure}
\includegraphics[width=0.45\textwidth]{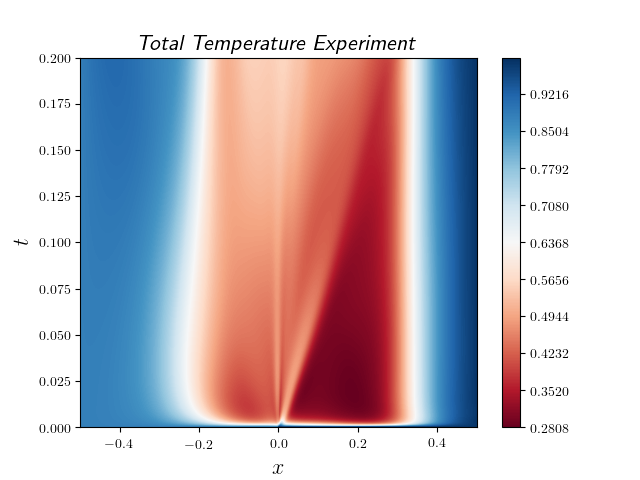} \includegraphics[width=0.45\textwidth]{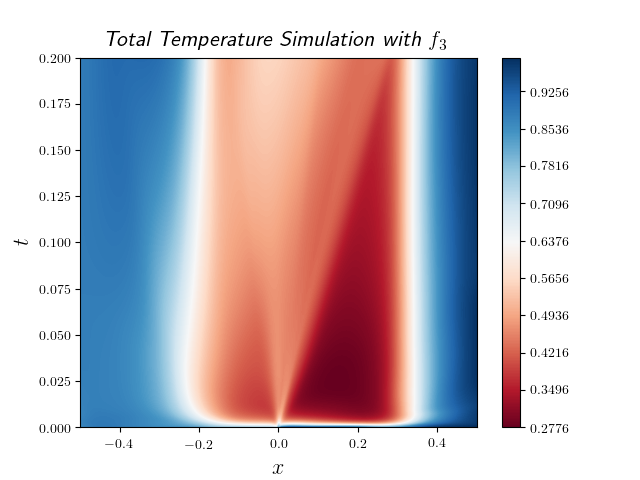}\\  \includegraphics[width=0.45\textwidth]{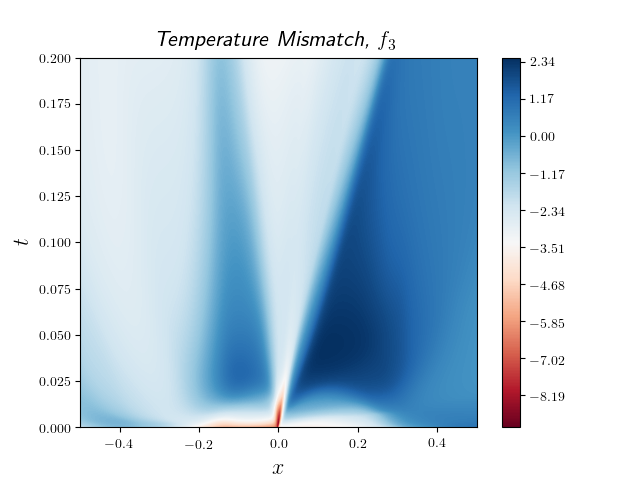} \includegraphics[width=0.45\textwidth]{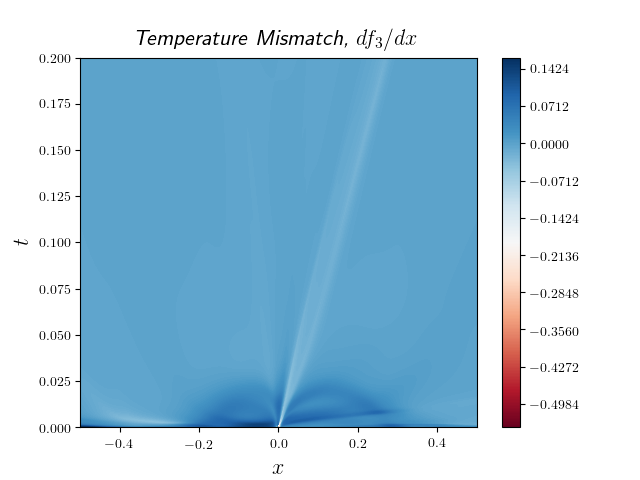} \caption{The experimental temperature profile is given on the top left, and the simulation on the top right with $f_3$.  The mismatch function $f_3$ is shown on the bottom left, with its $x$-derivative on the bottom right.}\label{fig:11}
\end{figure}

Given the power of the DNN framework, to reveal the mismatch of the underlying system that is actually observed in the experiment, DNN is trained to model the following enriched PDE system instead of (\ref{class}): \begin{equation}\begin{aligned}\label{enrich} \partial_{t}\rho & + (\rho u)_x = f_1, \\  \partial_{t}(\rho u) & + \left(\rho u^2 + p -\tau\rho u_x\right)_x = f_2, \\ \partial_{t}E & + \left(Eu+pu-\tau\rho u u_x-\kappa T_{x} \right)_x = f_3 \end{aligned}\end{equation} where \[E = \frac{p}{\gamma -1} + \frac{\rho}{2}u^2,\]  with again the same initial-boundary data from Table~\ref{table:1}.

The solution from the experiment (\ref{exp}) is now used as training data to supervise (\ref{enrich}).   Formally the objective function from (\ref{loss}) is enriched with the training data, $\boldsymbol{U}_{\mathrm{exp}} = (\rho_{\mathrm{exp}} , \rho_{\mathrm{exp}}, u_{\mathrm{exp}}, E_{\mathrm{exp}})$ from (\ref{exp}), by appending the new loss functions \[ \|\rho_{\mathrm{exp}} -\rho\|_{\mathfrak{L}_s(\Omega\times [0,T_{s}],\mathscr{E})} +  \|\rho_{\mathrm{exp}} u_{\mathrm{exp}} - \rho u\|_{\mathfrak{L}_s(\Omega\times [0,T_{s}],\mathscr{E})} + \| E_{\mathrm{exp}} - E\|_{\mathfrak{L}_2(\Omega\times [0,T_{s}],\mathscr{E})}\] to (\ref{loss}), where $\mathscr{E}$ are the experimentally measured data support points, and $\mathfrak{L}_{s}$ is the mean square error.   For simplicity, the support points $\mathscr{E}$ are chosen as a $1000\times1000$ point grid in $(x,t)$.  In addition, the neural network outputs associated to the $f_i=f_i(x,t)$ are $L^2$-regularized by setting: \[\sum_{i}\|f_i^{2}\|_{\mathfrak{L}_i(\Omega\times [0,T_s], \wp_{i})},\] where here the distributions $\wp_i$ are chosen to be the same as for (\ref{enrich}), and the $\mathfrak{L}_i$ are $L^2$-losses with weight $w_i=0.0001$ for each $i$ in order to minimize the penalization for accumulating non-zero $f_i$.  

\begin{figure}
\includegraphics[width=0.45\textwidth]{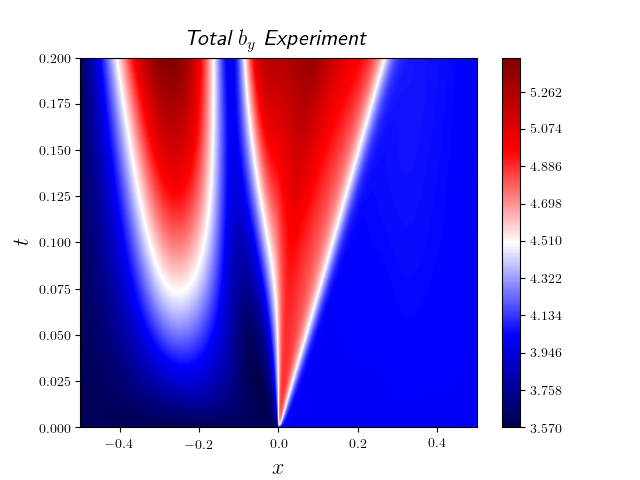} \includegraphics[width=0.45\textwidth]{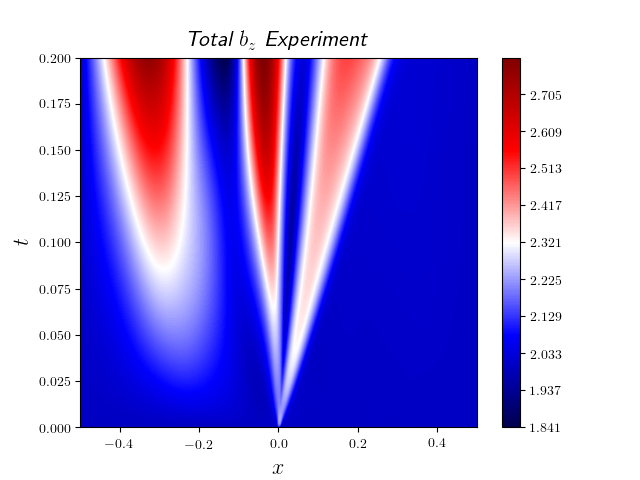} \caption{The experimental magnetic fields, where $b_x$ is measured to be negligibly small.}\label{fig:12}
\end{figure}

The resulting system matches up reasonably well to the experimental data.  As can be seen in Fig.~\ref{fig:9}, the simulation density and experimental density match to within the eyeball norm, where the initial mismatch is corrected with the $f_1$ function.  Though the total density nearly completely obfuscates the oscillations from the reactive subsystem in Fig.~\ref{fig:8}, the total density is off primarily near the initial state, where the reactive subsystem indicate a dramatic influx of mass due to the species being tracked by the sensor data in the experiment.  This type of mismatch in the mass conservation clearly indicates chemical reactions relative to the tracked species densities, unless, of course, the system is somehow not closed.  In Fig.~\ref{fig:10} the velocity mismatch is similarly convincing, where the mismatch $f_2$ shows complicated behavior with nontrivial variation that cannot be readily explained by the mass influx in $f_1$.  Similarly the mismatch $f_3$ in the temperature in Fig.~\ref{fig:11} also shows an unusual signature not present in $f_1$.

Returning to our list of potential questions about how the model system relates to the experimental system (\ref{exp}), we can now make some fairly strong conclusions.  First, clearly there is a mass mismatch in the measured species density.  The mismatch is substantial and cannot be captured by simple parameter estimation.  Therefore, there is physics in the mass equation that is not being accounted for by the ascribed model system (\ref{class}).   Moreover, it is also straightforward to conclude that the giant influx of mass makes a chemical reaction a very likely candidate to explain the experimental behavior, assuming the system closed.  

In contrast to the mass equation, the momentum and energy equations in (\ref{class}) show perturbations away from the model Sod shock solution that indicate that complicated system dynamics is occurring in the continuum.  While this behavior might be more challenging to diagnose, it is clear in these cases too that the base equations (\ref{class}), even with the addition of $f_1$, cannot account for the system response, and there is missing physics.  As it so happens, knowing simply the magnetic field variation of the experiment, as shown in Fig.~\ref{fig:12} is really enough to diagnose the mismatch in both the the velocity $f_2$ and temperature $f_3$ as related to the magnetic field, since they exhibit signature features that track the magnetic field variation.
    
\section{Conclusion}

Gridless representations provided by deep neural networks in combination with numerical optimization can be used to solve complicated systems of differential equations that display irregular solutions. This requires fairly straightforward techniques, and in irregular solutions benefits from the usual trick of adding numerical diffusion to smooth numerically unstable function representations.  The DNN method compares favorably with regard to accuracy and stability to more conventional numerical methods and yet enables one-shot exploration of the entire parameter space.  The incorporation of efficient optimization algorithms in DNN methods into the PDE solver lends itself to an ease and simplicity of integrating advanced data analytic techniques into physics-enriched model systems.  Early results indicate that DNNs enable a simple and powerful framework for exploring and advancing predictive capabilities in science and engineering.

\section{Acknowledgements}

This work was supported by U.S. DOE Contract No. DE-FG02-04ER54742 and U.S. DOE Office of Fusion Energy Sciences Scientific Discovery through Advanced Computing (SciDAC) program under Award Number DE-SC0018429.  This work was also supported by the NSF grant AST-1413501.  We acknowledge the Texas Advanced Computing Center at The University of Texas at Austin for providing HPC resources. Computations were performed on the Maverick2 GPU cluster.

\bibliography{cem.bib}
\bibliographystyle{ieeetr}

\end{document}